\documentclass[sigconf]{acmart}

\usepackage{caption}
\usepackage{filecontents}
\usepackage{url}
\usepackage{amsmath}
\usepackage{graphicx}
\usepackage{url}
\usepackage{hyperref}
\usepackage{amssymb,enumitem}
\usepackage{amsfonts,amssymb}
\usepackage{color, soul}
\usepackage{mathrsfs}
\usepackage{cleveref}
\usepackage{multirow}
\usepackage{float}
\usepackage{wrapfig}
\usepackage{amsthm}
\usepackage{algorithm}
\usepackage{algorithmic}
\usepackage{bm}
\usepackage{booktabs}
\usepackage{enumitem}
\usepackage{mathtools}
\newtheorem{myDef}{Definition}
\newtheorem{myDef1}{Problem}
\newtheorem{proposition}{Proposition}
\newtheorem{theorem}{Theorem}
\newtheorem{lemma}{Lemma}
\newtheorem{assumption}{Assumption}

\usepackage{epstopdf}
\newtheorem{appendix_proposition}{Proposition}
\newtheorem{appendix_theorem}{Theorem}
\newtheorem{appendix_lemma}{Lemma}
\newtheorem{appendix_assumption}{Assumption}

\usepackage{tikz}
\usepackage{subcaption}
\usepackage{pgfplots}
\pgfplotsset{compat=1.14}
\usetikzlibrary{arrows,positioning,automata,calc,shapes}
\pgfplotsset{compat=newest, scaled z ticks=false} 
\pgfplotsset{plot coordinates/math parser=false}
\newlength\figureheight 
\newlength\figurewidth

\AtBeginDocument{%
  \providecommand\BibTeX{{%
    \normalfont B\kern-0.5em{\scshape i\kern-0.25em b}\kern-0.8em\TeX}}}

\copyrightyear{2024}
\acmYear{2024}
\setcopyright{rightsretained}
\acmConference[KDD '24]{Proceedings of the 30th ACM SIGKDD Conference on Knowledge Discovery and Data Mining}{August 25--29, 2024}{Barcelona, Spain}
\acmBooktitle{Proceedings of the 30th ACM SIGKDD Conference on Knowledge Discovery and Data Mining (KDD '24), August 25--29, 2024, Barcelona, Spain}\acmDOI{10.1145/3637528.3671744}
\acmISBN{979-8-4007-0490-1/24/08}

\makeatletter
\gdef\@copyrightpermission{
  \begin{minipage}{0.3\columnwidth}
   \href{https://creativecommons.org/licenses/by/4.0/}{\includegraphics[width=0.90\textwidth]{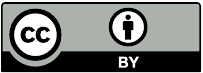}}
  \end{minipage}\hfill
  \begin{minipage}{0.7\columnwidth}
   \href{https://creativecommons.org/licenses/by/4.0/}{This work is licensed under a Creative Commons Attribution International 4.0 License.}
  \end{minipage}
  \vspace{5pt}
}
\makeatother

\begin{document}

\title{IDEA: A Flexible Framework of Certified Unlearning for \\ Graph Neural Networks}




\author{Yushun Dong}
\affiliation{%
  \institution{The University of Virginia}
  \city{Charlottesville}
    \country{USA}
}
\email{yd6eb@virginia.edu}

\author{Binchi Zhang}
\affiliation{%
  \institution{The University of Virginia}
  \city{Charlottesville}
    \country{USA}
}
  \email{epb6gw@virginia.edu}

\author{Zhenyu Lei}
\affiliation{%
  \institution{The University of Virginia}
  \city{Charlottesville}
    \country{USA}
}
  \email{vjd5zr@virginia.edu}

\author{Na Zou}
\affiliation{%
  \institution{The University of Houston}
  \city{Houston}
    \country{USA}
}
  \email{nzou2@uh.edu}

\author{Jundong Li}
\affiliation{%
  \institution{The University of Virginia}
  \city{Charlottesville}
    \country{USA}
}
  \email{jundong@virginia.edu}








\renewcommand{\shortauthors}{Yushun Dong, Binchi Zhang, Zhenyu Lei, Na Zou, \& Jundong Li}

\begin{abstract}


Graph Neural Networks (GNNs) have been increasingly deployed in a plethora of applications. However, the graph data used for training may contain sensitive personal information of the involved individuals. Once trained, GNNs typically encode such information in their learnable parameters. As a consequence, privacy leakage may happen when the trained GNNs are deployed and exposed to potential attackers. 
Facing such a threat, machine unlearning for GNNs has become an emerging technique that aims to remove certain personal information from a trained GNN. Among these techniques, certified unlearning stands out, as it provides a solid theoretical guarantee of the information removal effectiveness. Nevertheless, most of the existing certified unlearning methods for GNNs are only designed to handle node and edge unlearning requests. Meanwhile, these approaches are usually tailored for either a specific design of GNN or a specially designed training objective. These disadvantages significantly jeopardize their flexibility. In this paper, we propose a principled framework named IDEA to achieve flexible and certified unlearning for GNNs. Specifically, we first instantiate four types of unlearning requests on graphs, and then we propose an approximation approach to flexibly handle these unlearning requests over diverse GNNs. We further provide theoretical guarantee of the effectiveness for the proposed approach as a certification. Different from existing alternatives, IDEA is not designed for any specific GNNs or optimization objectives to perform certified unlearning, and thus can be easily generalized. Extensive experiments on real-world datasets demonstrate the superiority of IDEA in multiple key perspectives.
\end{abstract}






\maketitle

\section{Introduction}

Graph-structured data is ubiquitous among various real-world applications, such as online social platform~\cite{hamilton2017inductive}, finance system~\cite{wang2019semi}, and chemical discovery~\cite{irwin2012zinc}. In recent years, Graph Neural Networks (GNNs) have exhibited promising performance in various graph-based downstream tasks~\cite{hamilton2017inductive,zhou2020graph,wu2020comprehensive,xu2018powerful}. The success of GNNs is mainly attributed to its message-passing mechanism, which enables each node to take advantage of the information from its multi-hop neighbors~\cite{hamilton2017inductive,kipf2016semi}. Therefore, GNNs have been widely adopted in a plethora of realms~\cite{zitnik2018modeling,do2019graph,zheng2020gman,fan2019graph,wu2019session,wang2022improving}.

Despite the success of GNNs, their widespread usage has also raised social concerns about the issue of privacy protection~\cite{wu2023certified,chien2022efficient,zheng2023graph}.
It is worth noting that, in practice, the graph data used for training may contain sensitive personal information of the involved individuals~\cite{wu2023certified,olatunji2021membership,wu2021adapting}. Once trained, these GNNs typically encode such personal information in the learnable parameters. As a consequence, privacy leakage may happen when the trained GNNs are deployed and exposed to potential attackers~\cite{olatunji2021membership,wu2021adapting}. 
%
%
For example, the similarity of the health records between patients could provide key information for disease diagnosis~\cite{zhang2021graph}. Therefore, GNNs can be trained on patient networks for disease prediction, where the connections between patients indicate high similarity scores of their health records. However, malicious attackers can easily reveal the patients' health records that are used for training via membership inference attack~\cite{olatunji2021membership}, which severely threatens privacy.
%
%
%
Facing such a threat of privacy leakage, legislation such as the General Data Protection Regulation (GDPR) (GDPR 2016)~\cite{regulation2018general}, the California Consumer Privacy Act (CCPA) (CCPA 2018)~\cite{pardau2018california}, and the Personal Information Protection and Electronic Documents Act (PIPEDA 2000)~\cite{act2000personal} have emphasized the importance of \textit{the right to be forgotten}~\cite{kwak2017let}. Specifically, users should have the right to request the deletion of their personal information from those learning models that encode it. Such an urgent need poses challenges towards removing certain personal information from the trained GNNs.

The need for information removal from these trained models has led to the development of \textit{machine unlearning}~\cite{bourtoule2021machine,xu2023machine}.
Specifically, the ultimate goal of machine unlearning is to remove information regarding certain training data from a previously trained model.
A straightforward approach is to perform model re-training. However, on the one hand, the model owner may not have full access to the training data; on the other hand, re-training can be prohibitively expensive even if training data is fully accessible~\cite{duan2022comprehensive}. 
To achieve more efficient information removal, a series of existing works~\cite{bourtoule2021machine,cao2015towards,izzo2021approximate} proposed to directly modify the parameters of the trained models. Nevertheless, most of these works only achieve unlearning empirically and fail to provide any theoretical guarantee. This problem has led to the emerging of certified unlearning~\cite{sekhari2021remember,guo2019certified}, which aims to develop unlearning approaches with theoretical guarantee on their effectiveness. In the domain of graph learning, a few recent works, such as~\cite{wu2023certified,chien2022efficient}, have explored to achieve certified unlearning for GNNs. However, a major limitation of these approaches is their low flexibility. 
First, most approaches are designed to completely unlearn a given set of nodes or edges, while this may not comply with certain unlearning needs in real-world applications. For example, on a social network platform, a user may decide to stop disclosing certain personal information to the GNN-based friend recommendation model but continue using the platform. In such a case, the attribute information of this user should then be partially removed from the GNN model, which protects the user's privacy and maintains algorithmic personalization as well.
Therefore, it is desired to develop flexible certified unlearning approaches for GNNs to handle unlearning requests centered on node attributes.
Second, existing certified unlearning approaches are mostly designed for a specific type of GNNs~\cite{chien2022efficient} or the GNNs trained following a specially designed objective function~\cite{wu2023certified,chien2022efficient}. However, various GNNs and objectives have been adopted for diverse real-world applications, and thus it is also desired to develop more flexible certified unlearning approaches for different GNNs trained with different objectives.
Nevertheless, existing exploration in developing flexible and certified unlearning approaches for GNNs remains nascent.

In this paper, we study a novel and critical problem of developing a certifiable unlearning framework that can flexibly unlearn personal information in graphs and generalize across GNNs. We note that this is a non-trivial task. In essence, we mainly face three challenges. \textit{(i) Characterizing node dependencies.} Different from tabular data, the nodes in graph data usually have dependencies with each other. Properly characterizing node dependencies thus becomes the first challenge to achieve unlearning for GNNs.
\textit{(ii) Achieving flexible unlearning.} Unlearning requests may be initiated towards nodes, node attributes (partial or full), and edges. Meanwhile, various GNNs have been adopted for different applications, and most of these GNNs have different model structures and optimization objectives. Therefore, achieving flexible unlearning for different types of unlearning requests, GNN structures, and objectives becomes the second challenge.
\textit{(iii) Obtaining certification for unlearning.} To reduce the risk of privacy leakage, it is critical for the model owner to ensure that the information needed to be removed has been completely wiped out before model deployment. However, GNNs may have complex structures, and it is difficult to examine whether certain sensitive personal information remains being encoded or not.
Meanwhile, certified unlearning for GNNs usually requires strict conditions (e.g., assuming that GNNs are trained under a specially designed objective~\cite{wu2023certified,chien2022efficient}) and thus sacrifices flexibility.
Properly certifying the effectiveness of unlearning is our third challenge.


\textbf{Our Contributions.} We propose IDEA (flex\underline{I}ble an\underline{D} c\underline{E}rtified unle\underline{A}rning), which is a flexible framework of certified unlearning for GNNs. Specifically, to tackle the first two challenges, we propose to model the intermediate state between the optimization objectives with and without the instances (e.g., nodes, edges, and attributes) to be unlearned. Meanwhile, four different types of common unlearning requests are instantiated, and GNN parameters after unlearning can be efficiently approximated with flexible unlearning request specifications.
To tackle the third challenge, we propose a novel theoretical certification on the unlearning effectiveness of IDEA. We show that our certification method brings an empirically tighter bound on the distance between the approximated and actual GNN parameters compared to other existing alternatives. We summarize our contributions as: \textbf{(1) Problem Formulation.} We formulate and make an initial investigation on a novel research problem of flexible and certified unlearning for GNNs. \textbf{(2) Algorithm Design.} We propose IDEA, a flexible framework of certified unlearning for GNNs without relying on any specific GNN structures or any specially designed objective functions, which shows significant value for practical use. \textbf{(3) Experimental Evaluation.} We conduct comprehensive experiments on real-world datasets to verify the superiority of IDEA over existing alternatives in multiple key perspectives, including bound tightness, unlearning efficiency, model utility, and unlearning effectiveness.
    


\section{Preliminaries}


\subsection{Notations}


We use bold uppercase letters (e.g., $\bm{A}$), bold lowercase letters (e.g., $\bm{x}$), and normal lowercase letters (e.g., $n$) to denote matrices, vectors, and scalars, respectively. 
We represent an attributed graph as $\mathcal{G} = \{\mathcal{V}, \mathcal{E}, \mathcal{X}\}$. Here $\mathcal{V} = \{v_1, ..., v_n\}$ denotes the set of nodes, where $n$ is the total number of nodes. $\mathcal{E} \subset \mathcal{V} \times \mathcal{V}$ represents the set of edges. $\mathcal{X} = \{x_{1,1}, ..., x_{n,c}\}$ is the set of node attribute values, where $c$ is the total number of node attribute dimensions, and $x_{i,j}$ represents the attribute value of node $v_i$ at the $j$-th attribute dimension ($1 \leq i \leq n, 1 \leq j \leq c$).
%
%
%
We utilize $f_{\bm{\theta}}$ to represent a GNN model parameterized by the learnable parameters in $\bm{\theta}$.

In this paper, we focus on the commonly studied
node classification task, which widely exists
in real-world applications. Specifically, we are given the labels of a set of training nodes $\mathcal{V}_{\text{trn}}$ ($\mathcal{V}_{\text{trn}} \subset \mathcal{V}$) as $\mathcal{Y}_{\text{trn}}$.
Here $\mathcal{Y}_{\text{trn}} = \{Y_1, ..., Y_m\}$, where $Y_i \in \{1, ..., c\}$ ($ 1 \leq i \leq m$) is the node label of $v_i$; $c$ is the total number of possible classes; and $m$ represents the number of training nodes, i.e., $m = |\mathcal{V}_{\text{trn}}|$.
Our goal here is to optimize the parameter $\bm{\theta}$ of the GNN model $f$ with $k$ message-passing layers as $\bm{\theta}^*$ w.r.t. certain objective function over $\mathcal{V}_{\text{trn}}$, such that $f_{\bm{\theta}^*}$ is able to achieve accurate predictions for the nodes in the test set $\mathcal{V}_{\text{tst}}$ ($\mathcal{V}_{\text{tst}} \cap \mathcal{V}_{\text{trn}} = \varnothing$).

\subsection{Problem Statement}




In this subsection, we formally present the problem formulation of \textit{Flexible and Certified Unlearning for GNNs}. We first elaborate on the mathematical formulation of certified unlearning for GNNs.
%
Specifically, certified unlearning requires that the unlearning strategy have a theoretical guarantee of unlearning effectiveness.
We adopt a commonly used criterion for the effectiveness of unlearning, i.e., \textit{$(\varepsilon-\delta)$ Certified Unlearning}. Here $\varepsilon$ and $\delta$ are two parameters controlling the relaxation of such a criterion.
We present the definition of $(\varepsilon-\delta)$ certified unlearning for GNNs below.
%
%
\begin{myDef}
\label{unlearning4}
\textbf{$(\varepsilon-\delta)$ Certified Unlearning for GNNs.} Let $\mathcal{H}$ be the hypothesis space of a GNN model parameters and $\mathcal{A}$ be the associated optimization process. Given a graph $\mathcal{G}$ for GNN optimization and a $\Delta \mathcal{G}$ that characterizes the information to be unlearned, $\mathcal{U}$ is an ($\varepsilon-\delta$) certified unlearning process iff $\forall\;\mathcal{T}\subseteq\mathcal{H}$, we have 
\begin{align}
\mathrm{Pr}\left(\mathcal{U}\left(\mathcal{G},\Delta\mathcal{G},\mathcal{A}\left(\mathcal{G}\right)\right)\in\mathcal{T}\right)\leq e^{\varepsilon}\mathrm{Pr}\left(\mathcal{A}\left(\mathcal{G} \ominus \Delta\mathcal{G} \right)\in\mathcal{T}\right)+\delta \text{, and } \notag \\
\mathrm{Pr}\left(\mathcal{A}\left(\mathcal{G}\ominus \Delta\mathcal{G}\right)\in\mathcal{T}\right)\leq e^{\varepsilon}\mathrm{Pr}\left(\mathcal{U}\left(\mathcal{G},\Delta\mathcal{G},\mathcal{A}\left(\mathcal{G}\right)\right)\in\mathcal{T}\right)+\delta, \notag
\end{align}
where $\mathcal{G} \ominus \Delta\mathcal{G}$ represents the graph data with the information characterized by $\Delta\mathcal{G}$ being removed.
\end{myDef}
The intuition of Definition~\ref{unlearning4} is that, once the two inequalities above are satisfied, the difference between the distribution of the unlearned GNN parameters and that of the re-trained GNN parameters over $\mathcal{G} \ominus \Delta\mathcal{G}$ is bounded by a small threshold $\varepsilon$ and relaxed by a probability $\delta$. 
%
We note that, different from most existing literature on GNN unlearning, the information to be unlearned does not necessarily come from a node or an edge in Definition~\ref{unlearning4}.
Such an extension paves the way towards more flexible certified unlearning for GNNs.
%
%
We now formally present the problem formulation of \textit{Flexible and Certified Unlearning for GNNs} below.



\begin{myDef1}
\label{p1}
\textbf{Flexible and Certified Unlearning for GNNs.} Given a GNN model $f_{\bm{\theta}^*}$ optimized over $\mathcal{G}$ and any request to unlearn information characterized by $\Delta \mathcal{G}$, our goal is to achieve $(\varepsilon-\delta)$ certified unlearning over $f_{\bm{\theta}^*}$.
\end{myDef1}

\section{Unlearning Request Instantiations}
\label{request_instantiations}

\begin{figure}[t]
    \centering
    \vspace{1mm}
\includegraphics[width=0.99\linewidth]{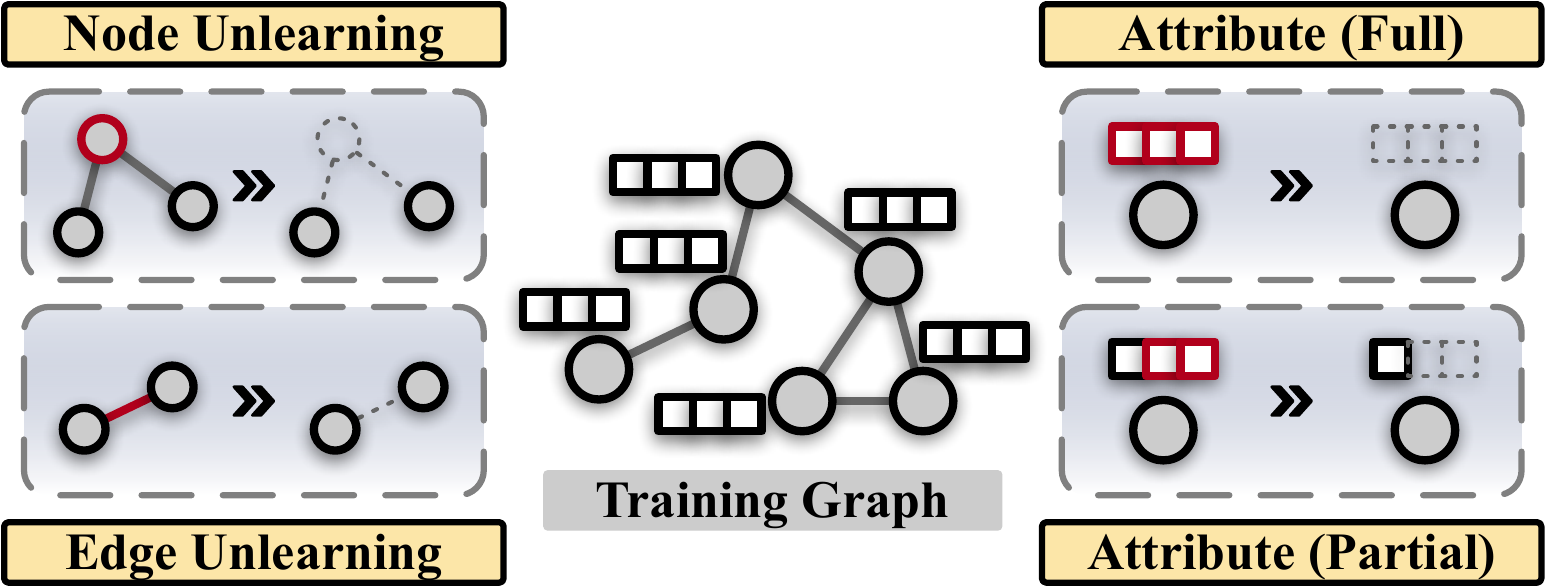}
    \caption{An illustration of common unlearning requests.}
    \label{requests_figure}
    \vspace{-6mm}
\end{figure}

We instantiate the unlearning requests characterized by $\Delta \mathcal{G}$, namely \textit{Node Unlearning Request}, \textit{Edge Unlearning Request}, and \textit{Attribute Unlearning Request}. We present an illustration in Figure~\ref{requests_figure}.

\noindent \textbf{Node Unlearning Request.} The most common unlearning request in GNN applications is to unlearn a given set of nodes. For example, in a social network platform, a GNN model can be trained on the friendship network formed by the platform users to perform friendship recommendation. When a user has decided to quit such a platform and withdrawn the consent of using her private data, this user may request to unlearn the node associated with her from the social network.
In such a case, the information to be unlearned is characterized by $\Delta \mathcal{G} = \{\Delta \mathcal{V}, \kappa_e(\Delta \mathcal{V}), \kappa_x(\Delta \mathcal{V})\}$. Here $\kappa_e$ and $\kappa_x$ return the set of the direct edges and node attributes associated with nodes in $\Delta \mathcal{V}$, respectively.
%




\noindent \textbf{Edge Unlearning Request.} 
In addition to the information encoded by the nodes, edges can also encode critical private information and may need to be unlearned as well.
In fact, it has been empirically proved that malicious attackers can easily infer the edges used for training, which directly threatens privacy~\cite{he2021stealing}.
In such a case, the information to be unlearned is characterized by $\Delta \mathcal{G} = \{\varnothing, \Delta \mathcal{E}, \varnothing\}$.

\noindent \textbf{Attribute Unlearning Request.} 
Both requests above fail to represent cases where only node attributes are requested to be unlearned. Here we show two common node attribute unlearning requests. 
\textit{(1) Full Attribute Unlearning.} In this case, all information regarding the attributes of a set of nodes is requested to be unlearned. For example, a social network platform user may withdraw the consent for the GNN-based friend recommendation algorithm to encode any of its attributes during training. In such a case, the information to be unlearned is characterized by $\Delta \mathcal{G} = \{\varnothing, \varnothing, \Delta \mathcal{X}\}$, where for node $v_i$, if $x_{i,j} \in \Delta \mathcal{X}$, then $\forall j \in \{1, ..., c\}, x_{i,j} \in \Delta \mathcal{X}$.
\textit{(2) Partial Attribute Unlearning.} The attributes of a node may also be requested to be partially unlearned. For example, in a social network, a user may withdraw the consent of using the information regarding certain attribute(s) due to various reasons, e.g., feeling being unfairly treated. However, this user may still continue using such a platform, and thus other attributes should not be unlearned to ensure satisfying personalized service quality. In such a case, the information to be unlearned is characterized by $\Delta \mathcal{G} = \{\varnothing, \varnothing, \Delta \mathcal{X}\}$, where for node $v_i$, if $x_{i,j} \in \Delta \mathcal{X}$, then $\exists j \in \{1, ..., c\}, x_{i,j} \notin \Delta \mathcal{X}$.
%
%
%
Note that the two types of attribute unlearning can be requested together. Hence, we utilize $\Delta \mathcal{X}$ to characterize a mixture of both types of attributes.

Based on the instantiations above, we denote $\Delta \mathcal{G} = \{\Delta \mathcal{V}, \Delta \mathcal{E} \cup \kappa_e(\Delta \mathcal{V}), \Delta \mathcal{X} \cup \kappa_x(\Delta \mathcal{V})\}$ as a potential combination of all types of unlearning requests. Accordingly, we formally define $\mathcal{G} \ominus \Delta\mathcal{G} = \{\mathcal{V} \backslash \Delta \mathcal{V}, \mathcal{E} \backslash \Delta \mathcal{E} \backslash \kappa_e(\Delta \mathcal{V}), \mathcal{X} \backslash \Delta \mathcal{X} \backslash \kappa_x(\Delta \mathcal{V})\}$.


\section{Methodology}

In this section, we present our proposed framework IDEA, which aims to achieve flexible and certified unlearning for GNNs. We first present the general formulation of flexible unlearning for GNNs. Then, we introduce a unified modeling integrating different instantiations of unlearning requests. We finally propose a novel theoretical guarantee on the effectiveness of IDEA as the certification.

\subsection{Flexible Unlearning for GNNs}

We first present a unified formulation of flexible unlearning for GNNs. In general, our rationale here is to design a framework to directly approximate the change in the (optimal) learnable parameter $\bm{\theta}^*$ during unlearning.
Specifically, we first review the training process of a given GNN model $f$ over graph data $\mathcal{G}$. Then, we consider the training objective with information of $\Delta \mathcal{G}$ being removed as a perturbed training objective over $\mathcal{G}$. We are now able to analyze how the optimal learnable parameter $\bm{\theta}^*$ would change when the objective function is modified. Note that we adopt a generalized formulation of such modification over the objective function, such that our analysis can be adapted to different unlearning requests.

In a typical training process of a given GNN model $f$ over graph data $\mathcal{G}$, the optimal learnable parameter $\bm{\theta}^*$ is obtained via solving the optimization problem of
\begin{align}
\label{old_obj}
\arg \min _{\bm{\theta}} \frac{1}{m} \sum_{v_i \in \mathcal{V}_{\text{trn}}} \mathscr{L}\left(\bm{\theta}, v_i, \mathcal{G}\right),
\end{align}
where a typical choice of $\mathscr{L}$ is cross-entropy loss in node classification tasks. Here we consider that the computation of $\mathscr{L}$ also relies on other necessary information such as $\hat{Y}_i$ by default and omit them for simplicity.
As a comparison, the optimal learnable parameter trained over $\mathcal{G} \ominus \Delta\mathcal{G}$, which we denoted as $\tilde{\bm{\theta}}^*$, is obtained via solving the problem of
\begin{align} 
\label{new_obj}
\arg \min _{\bm{\theta}} \frac{1}{m - |\Delta\mathcal{V}|} \sum_{v_i \in \mathcal{V}_{\text{trn}}\backslash\Delta\mathcal{V}} \mathscr{L}\left(\bm{\theta}, v_i, \mathcal{G} \ominus \Delta\mathcal{G}\right).
\end{align}
To study how the optimal parameters change when transforming from Equation (\ref{old_obj}) to Equation (\ref{new_obj}), it is necessary to analyze how the objective function and optimal solution change between the two cases.
To systematically compare Equation (\ref{old_obj}) and Equation (\ref{new_obj}), here we define $\phi_k(\cdot)$ as a function that takes a node and a graph as its input and outputs the set of nodes in the computation graph of the input node (excluding the input node itself).
Here a computation graph is a subgraph centered on a given node with neighboring nodes up to $k$ hops away, where $k$ is the layer number of the studied GNN. Then we have the following proposition.

\begin{proposition}
\label{equivalence}
\textbf{Localized Equivalence of Training Nodes.} Given $\Delta \mathcal{G} = \{\Delta \mathcal{V}, \Delta \mathcal{E}, \Delta \mathcal{X}\}$ to be unlearned and an objective $\mathscr{L}$ computed over $f_{\bm{\theta}}$, $\mathscr{L}\left(\bm{\theta}, v_i, \mathcal{G}\right) = \mathscr{L}\left(\bm{\theta}, v_i, \mathcal{G} \ominus \Delta\mathcal{G}\right)$ holds $\forall v_i \notin \phi_k(v_j) \cup \{v_j\}, v_j \in \Delta \mathcal{V}  \cup \gamma_e(\Delta \mathcal{E}) \cup \gamma_x(\Delta \mathcal{X})$. Here $\gamma_e$ and $\gamma_x$ return the set of nodes that directly connect to the edges in $\mathcal{E}$ and that have associated attribute(s) in $\mathcal{X}$, respectively.
\end{proposition}
The intuition of Proposition~\ref{equivalence} is that, under a given $f_{\bm{\theta}}$, the value of $\mathscr{L}$ maintains the same between Equation (\ref{old_obj}) and Equation (\ref{new_obj}) for those training nodes that are not topologically close to the instances (i.e., nodes, attributes, and edges) in $\Delta \mathcal{G}$.
To bridge Equation (\ref{old_obj}) and Equation (\ref{new_obj}), we then propose a formulation to characterize the intermediate state.
Inspired by a series of previous works such as~\cite{wu2023gif,wu2023certified}, we add an additional term by defining $\bm{\theta}_{\Delta \mathcal{G}, \xi}^{*}$ with
%
%
\begin{align} 
\label{changed_obj}
\bm{\theta}_{\Delta \mathcal{G}, \xi}^{*} \coloneqq \arg \min _{\bm{\theta}} \frac{1}{m} \sum_{v_i \in \mathcal{V}_{\text{trn}}} \mathscr{L}\left(\bm{\theta}, v_i, \mathcal{G}\right) + \xi \left( \mathscr{L}_{\text{add}} - \mathscr{L}_{\text{sub}}\right).
\end{align}
We then introduce the modeling of $\mathscr{L}_{\text{add}}$ and $\mathscr{L}_{\text{sub}}$. Specifically, we propose to formulate $\mathscr{L}_{\text{add}}$ with 
\begin{align} 
\label{l_add}
\mathscr{L}_{\text{add}} &= \alpha_1 \sum_{v_i \in \mathcal{V}_1} \mathscr{L}\left(\bm{\theta}, v_i, \mathcal{G} \ominus \Delta\mathcal{G}\right) + \alpha_2 \sum_{v_i \in \mathcal{V}_2} \mathscr{L}\left(\bm{\theta}, v_i, \mathcal{G} \ominus \Delta\mathcal{G}\right) \notag \\ + \alpha_3 &\sum_{v_i \in \mathcal{V}_3} \mathscr{L}\left(\bm{\theta}, v_i, \mathcal{G} \ominus \Delta\mathcal{G}\right) + \alpha_4 \sum_{v_i \in \mathcal{V}_4} \mathscr{L}\left(\bm{\theta}, v_i, \mathcal{G} \ominus \Delta\mathcal{G}\right).
\end{align}
Here $\alpha_1, \alpha_2, \alpha_3, \alpha_4 \in \{0, 1\}$ are used to flag whether the requests of node unlearning, full attribute unlearning, partial node attribute unlearning, and edge unlearning exist or not, respectively. We now introduce $\mathcal{V}_1$, $\mathcal{V}_2$, $\mathcal{V}_3$, and $\mathcal{V}_4$.
Specifically, $\mathcal{V}_1$ represents the set of training nodes whose computation graph includes those nodes to be unlearned. We denote the sets of nodes associated with $\Delta \mathcal{X}$ when their unlearned attributes are replaced with any non-informative numbers (e.g., 0) as $\mathcal{V}_x^{\text{(Full)}}$ and $\mathcal{V}_x^{\text{(Partial)}}$ for full and partial attribute unlearning, respectively.
$\mathcal{V}_2$ and $\mathcal{V}_3$ include training nodes whose computation graph includes attributes to be unlearned fully and partially plus the nodes in $\mathcal{V}_x^{\text{(Full)}}$ and $\mathcal{V}_x^{\text{(Partial)}}$, respectively; $\mathcal{V}_4$ is the set of nodes whose computation graph includes those edges to be unlearned. Mathematically, we formulate $\mathcal{V}_1$, $\mathcal{V}_2$, $\mathcal{V}_3$, and $\mathcal{V}_4$ as
\begin{align}
\mathcal{V}_1 &= \cup_{v_i \in  \Delta \mathcal{V}} \left(\phi_k(v_i) \cap \mathcal{V}_{\text{trn}} \right), \\
\mathcal{V}_2 &= \mathcal{V}_x^{\text{(Full)}} \cup \{v_i: v_i \in \phi_k(v_j) \cap \mathcal{V}_{\text{trn}}, \; v_j \in \mathcal{V}_x^{\text{(Full)}}\}, \\
\mathcal{V}_3 &= \mathcal{V}_x^{\text{(Partial)}} \cup \{v_i: v_i \in \phi_k(v_j) \cap \mathcal{V}_{\text{trn}}, \; v_j \in \mathcal{V}_x^{\text{(Partial)}}\}, \\
\mathcal{V}_4 &= \cup_{v_i \in \gamma_e(\Delta \mathcal{E})} \left(\phi_k(v_i) \cap \mathcal{V}_{\text{trn}} \right)
%
\end{align}
We then formulate $\mathscr{L}_{\text{sub}}$ as
\begin{align} 
\label{l_sub}
\mathscr{L}_{\text{sub}} = \alpha_1 \sum_{v_i \in \tilde{\mathcal{V}}_1} \mathscr{L}\left(\bm{\theta}, v_i, \mathcal{G} \right) + \alpha_2 \sum_{v_i \in \tilde{\mathcal{V}}_2} \mathscr{L}\left(\bm{\theta}, v_i, \mathcal{G} \right) \notag \\ + \alpha_3 \sum_{v_i \in \tilde{\mathcal{V}}_3} \mathscr{L}\left(\bm{\theta}, v_i, \mathcal{G} \right) + \alpha_4 \sum_{v_i \in \tilde{\mathcal{V}}_4} \mathscr{L}\left(\bm{\theta}, v_i, \mathcal{G} \right),
\end{align}
where $\tilde{\mathcal{V}}_1$ includes all nodes in $\Delta \mathcal{V}$ and the training nodes within $k$ hops away from the nodes in $\Delta \mathcal{V}$;
We denote the sets of nodes associated with $\Delta \mathcal{X}$ with their vanilla attributes as $\tilde{\mathcal{V}}_x^{\text{(Full)}}$ and $\tilde{\mathcal{V}}_x^{\text{(Partial)}}$ for full and partial attribute unlearning, respectively.
$\tilde{\mathcal{V}}_2$ and $\tilde{\mathcal{V}}_3$ include training nodes whose computation graph includes attributes to be unlearned fully and partially plus the nodes in $\tilde{\mathcal{V}}_x^{\text{(Full)}}$ and $\tilde{\mathcal{V}}_x^{\text{(Partial)}}$, respectively;
$\tilde{\mathcal{V}}_4$ is the set of nodes whose computation graph includes those edges to be unlearned, i.e., $\tilde{\mathcal{V}}_4 = \mathcal{V}_4$. Mathematically, we have
\begin{align}
\tilde{\mathcal{V}}_1 &= \cup_{v_i \in  \Delta \mathcal{V}} \left(\phi_k(v_i) \cap \mathcal{V}_{\text{trn}} \right) \cup \Delta \mathcal{V}, \\
\tilde{\mathcal{V}}_2 &= \tilde{\mathcal{V}}_x^{\text{(Full)}} \cup \{v_i: v_i \in \phi_k(v_j) \cap \mathcal{V}_{\text{trn}}, \; v_j \in \tilde{\mathcal{V}}_x^{\text{(Full)}}\}, \\
\tilde{\mathcal{V}}_3 &= \tilde{\mathcal{V}}_x^{\text{(Partial)}} \cup \{v_i: v_i \in \phi_k(v_j) \cap \mathcal{V}_{\text{trn}}, \; v_j \in \tilde{\mathcal{V}}_x^{\text{(Partial)}}\}, \\
\tilde{\mathcal{V}}_4 &= \mathcal{V}_4 \label{last_modeling}.
\end{align}
We then have the complete formulation of Equation (\ref{changed_obj}) given Equation (\ref{l_add}) to (\ref{last_modeling}).
%
%
Based on the modeling above, we have the optimal equivalence between Equation (\ref{changed_obj}) and Equation (\ref{new_obj}) below.
\begin{lemma}
\label{new_equivalence}
\textbf{Optimal Equivalence.} The optimal solution to Equation (\ref{changed_obj}) (denoted as $\bm{\theta}_{\Delta \mathcal{G}, \xi}^{*}$) equals to the optimal solution to Equation (\ref{new_obj}) (denoted as $\tilde{\bm{\theta}}^{*}$) when $\xi = \frac{1}{m}$.
\end{lemma}

Now we have successfully bridged the gap between Equation (\ref{old_obj}) and Equation (\ref{new_obj}) by modeling their intermediate states with Equation (\ref{changed_obj}). More importantly, Lemma~\ref{new_equivalence} paves the way towards directly approximating $\tilde{\bm{\theta}}^*$ based on $\bm{\theta}^*$ by giving Theorem~\ref{influence} below.
\begin{theorem}
\label{influence}
\textbf{Approximation with Infinitesimal Residual.} Given a graph data $\mathcal{G}$, $\Delta \mathcal{G} = \{\Delta \mathcal{V}, \Delta \mathcal{E}, \Delta \mathcal{X}\}$ to be unlearned, and an objective $\mathscr{L}$ computed over an $f_{\bm{\theta}^*}$, using $\bm{\theta}^* + \frac{1}{m} \Delta \bar{\bm{\theta}}^*$ as an approximation of $\tilde{\bm{\theta}}^*$ only brings a first-order infinitesimal residual w.r.t. $\|\bm{\theta}^* - \tilde{\bm{\theta}}^*\|_2$, where $\Delta \bar{\bm{\theta}}^* = -\bm{H}^{-1}_{\bm{\theta}^*} \left( \nabla_{\bm{\theta}} \mathscr{L}_{\text{add}} - \nabla_{\bm{\theta}} \mathscr{L}_{\text{sub}}\right)$, and $\bm{H}_{\bm{\theta}^*} \coloneqq \nabla_{\bm{\theta}}^{2} \frac{1}{m} \sum_{v_i \in \mathcal{V}_{\text{trn}}} \mathscr{L}\left(\bm{\theta}, v_i, \mathcal{G}\right)$.
%
\end{theorem}
We note that the approximation strategy above relies on the assumption that $\forall \mathcal{V}_i \cap \mathcal{V}_j = \varnothing$ and $\forall \tilde{\mathcal{V}}_i \cap \tilde{\mathcal{V}}_j = \varnothing$ for $i, j \in \{1, 2, 3, 4\}$ when $i \neq j$. However, it can also handle cases where such an assumption does not hold. We show this in Proposition~\ref{sequential}.

\begin{proposition} 
\label{sequential}
    \textbf{Serializability of Approximation.} Any mixture of unlearning request instantiations can be split into multiple sets of unlearning requests, where each set of unlearning requests satisfies $\forall \mathcal{V}_i \cap \mathcal{V}_j = \varnothing$ and $\forall \tilde{\mathcal{V}}_i \cap \tilde{\mathcal{V}}_j = \varnothing$ for $i, j \in \{1, 2, 3, 4\}$ when $i \neq j$. Serially performing approximation following these request sets achieves upper-bounded error.
\end{proposition}

\noindent \textbf{Unlearning in Practice.} The approximation approach given by Theorem~\ref{influence} requires computing the inverse matrix of the Hessian matrix, which usually leads to high computational costs. Here we propose to utilize the stochastic estimation method~\cite{cook1980characterizations} to perform estimation based on an iterative approach, which reduces the time complexity to $O(t p)$. Here $t$ is the total number of iterations adopted by the stochastic estimation method, and $p$ represents the total number of learnable parameters in $\bm{\theta}$.

\subsection{Unlearning Certification}

In this subsection, we introduce a novel certification based on Theorem~\ref{influence}. According to the unlearning process given by Definition~\ref{unlearning4}, our goal is to achieve guaranteed closeness between $\tilde{\bm{\theta}}^*$ (i.e., the ideal unlearned parameter derived from Equation (\ref{new_obj})) and the approximation of such a parameter (denoted as $\bar{\bm{\theta}}^*$). In this way, we are able to achieve certifiable unlearning effectiveness.





\begin{figure}[t]
    \centering
\includegraphics[width=0.99\linewidth]{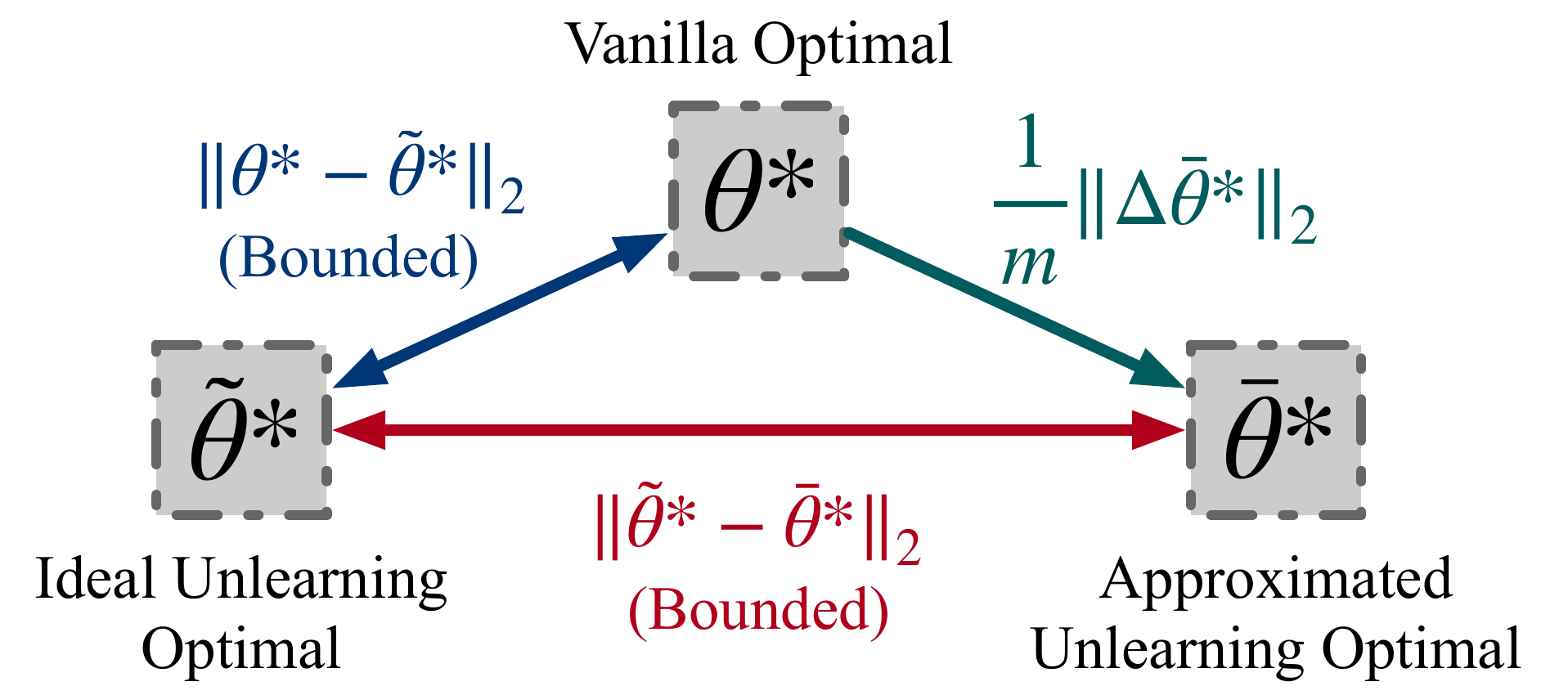}
    \caption{Distances between $\bm{\theta}^{*}$, $\tilde{\bm{\theta}}^*$, and $\bar{\bm{\theta}}^*$. Here, $\bm{\theta}^{*}$ denotes the optimal parameter before unlearning; $\tilde{\bm{\theta}}^*$ is the ideal optimal parameter after unlearning, which is obtained via re-training; $\bar{\bm{\theta}}^*$ is an approximation of $\tilde{\bm{\theta}}^*$ give by Theorem~\ref{influence}.}
    \label{certification_figure}
    \vspace{-2mm}
\end{figure}

Although certified unlearning for GNNs is studied by some recent explorations~\cite{wu2023certified,chien2022efficient}, these approaches can only be applied when the studied GNN model is trained following a specially modified objective. 
In particular, such a modification requires adding an additional regularization term of $\bm{\theta}$ scaled by a random vector onto the objective, which is specially designed for certification purposes.
However, most GNNs are optimized following common objectives (e.g., cross-entropy loss) instead of such a modified objective. Therefore, these certified unlearning approaches cannot be flexibly used across different GNNs in real-world applications.
Here we aim to develop a certified unlearning approach based on Theorem~\ref{influence}, such that it is not tailored for any optimization objective and thus can be easily generalized across various GNNs.
Towards this goal, we first review the $\ell_2$ distances between $\bm{\theta}^{*}$, $\tilde{\bm{\theta}}^*$, and $\bar{\bm{\theta}}^*$. We present an illustration in Figure~\ref{certification_figure}.
It is difficult to directly analyze the $\ell_2$ distance between $\tilde{\bm{\theta}}^*$ and $\bar{\bm{\theta}}^*$. We thus start by analyzing the $\ell_2$ distance between $\bm{\theta}^{*}$ and $\tilde{\bm{\theta}}^*$.
We found that the $\ell_2$ distance between $\bm{\theta}^{*}$ and $\tilde{\bm{\theta}}^*$ is upper bounded under common assumptions, which are widely adopted in other existing works tackling unlearning problems~\cite{wu2023certified,chien2022efficient,GuoGHM20}. We first present these assumptions below.

\begin{assumption} \label{assumption_main}
For the training objective of a given GNN model, we have: (1)
The loss values of optimal points are bounded: $|\mathscr{L}(\bm{\theta}^*)|\leq C$ and $|\mathscr{L}(\tilde{\bm{\theta}}^*)|\leq C$; (2) The loss function $\mathscr{L}$ is $L$-Lipschitz continuous; (3) The loss function $\mathscr{L}$ is $\lambda$-strongly convex.
\end{assumption}

Based on Assumption~\ref{assumption_main}, we now present the bound between $\bm{\theta}^{*}$ and $\tilde{\bm{\theta}}^*$ in Theorem~\ref{bound1}.

\begin{theorem}
\label{bound1}
\textbf{Distance Bound in Optimals.} The $\ell_2$ distance bound between $\tilde{\bm{\theta}}^*$ and $\bm{\theta}^*$ is given by 
\begin{align}
\|\tilde{\bm{\theta}}^*-\bm{\theta}^*\|_2 \leq\frac{L|\Delta\mathcal{V}|+\sqrt{4m\lambda C|\tilde{\mathcal{V}}|+L^2|\Delta\mathcal{V}|^2}}{m\lambda}.
\end{align}
Denote $\mathcal{V}_x^{\text{(F+P)}} = \mathcal{V}_x^{\text{(Full)}} \cup \mathcal{V}_x^{\text{(Partial)}}$, and $\tilde{\mathcal{V}}$ is given by
\begin{align}
\tilde{\mathcal{V}} = \mathcal{V}_1 \cup \mathcal{V}_4 \cup \{v_i: v_i \in \phi_k(v_j) \cap \mathcal{V}_{\text{trn}}, \; v_j \in \mathcal{V}_x^{\text{(F+P)}}\}.
\end{align}
\end{theorem}
Here the rationale of $\tilde{\mathcal{V}}$ is to describe the set of nodes whose computation graphs involve any instance (i.e., nodes, attributes, and edges) to be unlearned. Noticing the relationship between $\bm{\theta}^{*}$, $\tilde{\bm{\theta}}^*$, and $\bar{\bm{\theta}}^*$ give by Figure~\ref{certification_figure}, we further show the bound between $\tilde{\bm{\theta}}^*$ and $\bar{\bm{\theta}}^*$ in Proposition~\ref{final_triangle}.
\begin{proposition}
\label{final_triangle}
\textbf{Distance Bound in Approximation.}
The $\ell_2$ distance bound between $\tilde{\bm{\theta}}^*$ and $\bar{\bm{\theta}}^*$ is given by 
\begin{align}
\|\tilde{\bm{\theta}}^*-\bar{\bm{\theta}}^*\|_2 \leq\frac{\lambda \|\Delta \bar{\bm{\theta}}^*\|_2 + L|\Delta\mathcal{V}|+\sqrt{4m\lambda C|\tilde{\mathcal{V}}|+L^2|\Delta\mathcal{V}|^2}}{m\lambda}.
\end{align}
\end{proposition}
The rationale of Proposition~\ref{final_triangle} is to characterize the maximum $\ell_2$ distance between the ideal unlearning optimal and the approximation of unlearning optimal given by Theorem~\ref{influence}.
Finally, based on Proposition~\ref{final_triangle}, we are able to present the certification in Theorem~\ref{certification_last}.
\begin{theorem}
\label{certification_last}
Let $\bm{\theta}^* = \mathcal{A}\left(\mathcal{G}\right)$ be the empirical minimizer over $\mathcal{G}$, $\tilde{\bm{\theta}}^* = \mathcal{A}\left(\mathcal{G} \ominus \Delta \mathcal{G}\right)$ be the empirical minimizer over $\mathcal{G} \ominus \Delta \mathcal{G}$ and $\bar{\bm{\theta}}^*$ be an approximation of $\tilde{\bm{\theta}}^*$.
Define $\zeta$ as an upper bound of $\|\tilde{\bm{\theta}}^*-\bar{\bm{\theta}}^*\|_2$. We have
$\mathcal{U}\left(\mathcal{G},\Delta\mathcal{G},\mathcal{A}\left(\mathcal{G}\right)\right) = \bar{\bm{\theta}}^*+\bm{b}$ is an ($\varepsilon-\delta$) certified unlearning process, where $\bm{b}\sim\mathcal{N}(0,\sigma^2\bm{I})$ and $\sigma\geq\frac{\zeta}{\varepsilon}\sqrt{2\mathrm{ln(1.25/\delta)}}$.
\end{theorem}
Therefore, according to Theorem~\ref{certification_last}, we are able to achieve certified unlearning by adding zero-mean Gaussian noise over the approximation derived from Theorem~\ref{influence}.

\section{Experimental Evaluations}

We empirically evaluate the performance of IDEA in this section. In particular, we aim to answer the following research questions. \textbf{RQ1}: How tight can IDEA bound the $\ell_2$ distance between the ideal optimal $\tilde{\bm{\theta}}^*$ and the approximation $\bar{\bm{\theta}}^*$? \textbf{RQ2}: How well can IDEA improve the efficiency of unlearning compared with re-training and other alternatives? \textbf{RQ3}: How well can IDEA maintain the utility of the original GNN model? \textbf{RQ4}: How well can IDEA unlearn the information requested to be removed from the GNN?

\begin{figure*}[!t]
\centering
            \begin{subfigure}[t]{0.6\textwidth}
        \small
        \includegraphics[width=1.0\textwidth]{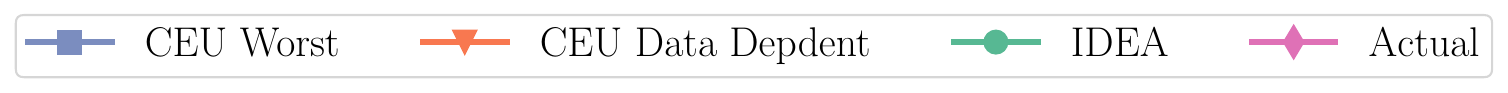}
        \end{subfigure} \\
        \begin{subfigure}[t]{0.33\textwidth}
        \small
        \centering
        \includegraphics[width=1.00\textwidth]{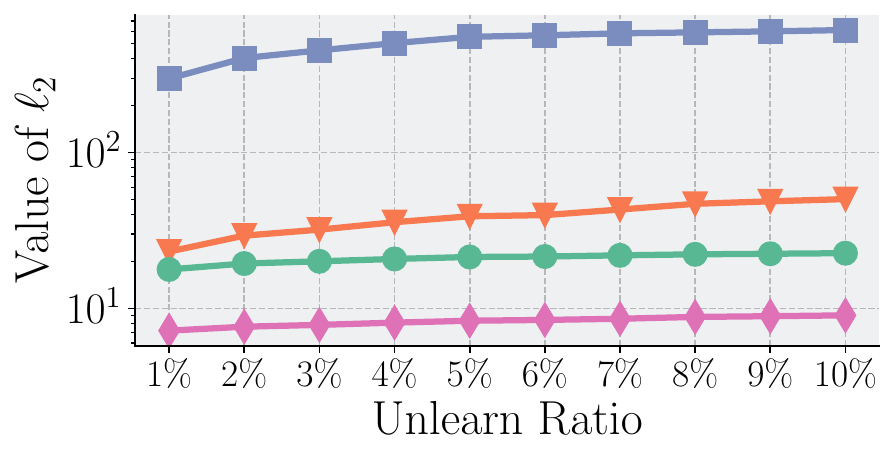}
        \vspace{-6mm}
            \caption[Network2]%
            {{Bounds vs. actual $\ell_2$ distance on Cora.}} 
            \label{cora_gcn}
        \end{subfigure}
            \begin{subfigure}[t]{0.33\textwidth}
        \small
        \centering
        \includegraphics[width=1.0\textwidth]{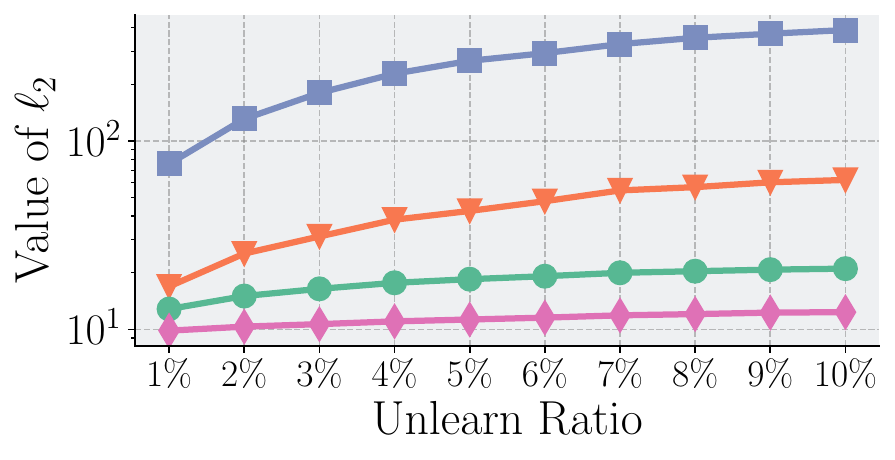}
        \vspace{-6mm}
            \caption[Network2]%
            {{Bounds vs. actual $\ell_2$ distance on Citeseer.}}   
            \label{citeseer_gcn}
        \end{subfigure}
            \begin{subfigure}[t]{0.33\textwidth}
        \small
        \centering
        \includegraphics[width=1.0\textwidth]{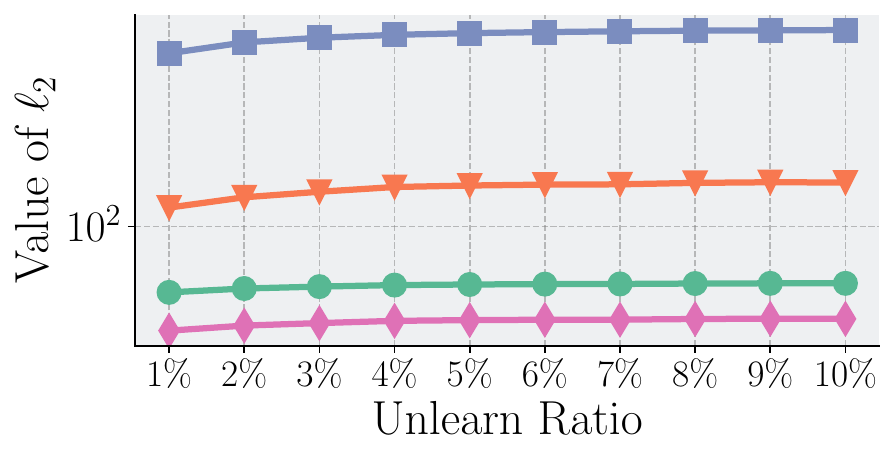}
        \vspace{-6mm}
            \caption[Network2]%
            {{Bounds vs. actual $\ell_2$ distance on PubMed.}}   
            \label{pubmed_gcn}
        \end{subfigure}
            \vspace{-2.5mm} 
    \caption{Bounds and actual value of the $\ell_2$ distance between $\tilde{\bm{\theta}}^*$ and $\bar{\bm{\theta}}^*$, i.e., $\|\tilde{\bm{\theta}}^*-\bar{\bm{\theta}}^*\|_2$, over Cora, CiteSeer and PubMed datasets. \textit{CEU Worst}, \textit{CEU Data Dependent}, \textit{IDEA}, and \textit{Actual} represent the worst bound based on CEU, the data-dependent bound based on CEU, the bound based on IDEA, and the actual value of $\|\tilde{\bm{\theta}}^*-\bar{\bm{\theta}}^*\|_2$ derived from re-training, respectively.}
    \label{gcn_bounds}
\end{figure*}

\subsection{Experimental Setup}

\noindent \textbf{Downstream Task and Datasets.} We adopt the widely studied node classification task as the downstream task, which accounts for a wide range of real-world applications based on GNNs. We perform experiments over five real-world datasets, including Cora~\cite{KipfW17}, Citeseer~\cite{KipfW17}, PubMed~\cite{KipfW17}, Coauthor-CS~\cite{shchur2018pitfalls,chen2022graph}, and Coauthor-Physics~\cite{shchur2018pitfalls,chen2022graph}. These datasets usually serve as commonly used benchmark datasets for GNN performance over node classification tasks. Specifically, Cora, Citeseer, and PubMed are citation networks, where nodes denote research publications and edges represent the citation relationship between any pair of publications. The node attributes are bag-of-words representations of the publication keywords. Coauthor-CS, and Coauthor-Physics are two coauthor networks, where nodes represent authors and edges denote the collaboration relationship between any pair of authors. We leave more dataset details, e.g., their statistics, in Appendix.



\noindent \textbf{Backbone GNNs.} To evaluate the generalization ability of IDEA across different GNNs, we propose to utilize two types of GNNs, including linear and non-linear GNNs. In terms of linear GNNs, we adopt the popular SGC~\cite{wu2019simplifying}; in terms of non-linear GNNs, we adopt three popular ones, including GCN~\cite{KipfW17}, GAT~\cite{velivckovic2017graph}, and GIN~\cite{XuHLJ19}.

\noindent \textbf{Unlearning Requests.} We consider all unlearning requests presented in Section~\ref{request_instantiations}. For each type of request, we perform experiments over a wide range of scales in terms of the number of unlearned instances (e.g., nodes and edges). For experiments with fixed ratios, we adopt a ratio of 5\% to perform unlearning for nodes or edges unless otherwise specified.

\noindent \textbf{Threat Models.} To evaluate the effectiveness of the unlearning strategy, we propose to adopt different types of threat models. Although IDEA is able to flexibly perform four different types of unlearning requests, there are only limited threat models can be chosen from. In our experiments, we adopt two state-of-the-art threat models, namely MIA-Graph~\cite{olatunji2021membership} and StealLink~\cite{he2021stealing}, for node membership inference attack and link stealing attack, respectively.

\noindent \textbf{Baselines.} We adopt five types of baselines for performance comparison. \textit{(1) Re-Training.} We adopt the re-training approach to obtain an ideal model based on the optimization problem given by Equation~(\ref{new_obj}). 
%
%
\textit{(2) Exact Unlearning.} We adopt the popular GraphEraser~\cite{chen2022graph} as a representative method for exact unlearning. Specifically, exact unlearning methods aim to achieve the exact same probability distribution in the model space (after unlearning) compared with the re-trained model. As a comparison, IDEA aims to approximate the distribution of the re-trained model through unlearning. 
%
%
\textit{(3) Certified Unlearning.} Finally, we adopt two representative approaches for certified unlearning, namely Certified Graph Unlearning (CGU)~\cite{chien2022certified} and Certified Edge Unlearning (CEU)~\cite{wu2023certified}. CGU is able to unlearn nodes, attributes, and edges. However, it is only applicable for the SGC model. As a comparison, CEU can be adapted to different GNNs. Nevertheless, it is specially designed for edge unlearning.

\noindent \textbf{Evaluation Metrics.} We evaluate IDEA with different metrics to answer the four research questions.
\textit{(1) Bound Tightness.} We propose to compare the numerical values of the bounds given by IDEA, the bounds given by other baselines, and the actual $\ell_2$ distance of model parameters yielded by re-training. A smaller bound on the $\ell_2$ distance indicates better tightness.
\textit{(2) Model Utility.} We utilize the F1 score to measure the model utility after unlearning. A higher F1 score indicates better performance.
\textit{(3) Unlearning Efficiency.} We utilize the running time (in seconds) that the unlearning methods take to measure efficiency, and a shorter running time indicates better efficiency.
\textit{(4) Unlearning Effectiveness.} We use the attack successful rate after unlearning to measure unlearning effectiveness. Lower attack successful rates indicate better effectiveness.


\begin{table*}[t]
\setlength{\tabcolsep}{15.65pt}
\renewcommand{\arraystretch}{0.8}
\centering
\caption{F1 score on five real-world graph datasets under node classification task. All numerical values are reported in percentage, and the F1 scores given by the proposed framework IDEA are marked in bold.}
\label{unlearning_utility}
\begin{tabular}{llccccc}
\toprule
                     &        & \textbf{Cora} & \textbf{CiteSeer} & \textbf{PubMed} & \textbf{CS} & \textbf{Physics} \\
                     \midrule
\multirow{5}{*}{\textbf{GCN}} & \textbf{Re-Training} & 76.88 $\pm$ 0.3     & 67.27 $\pm$ 0.6         & 76.20 $\pm$ 0.0       & 86.79 $\pm$ 0.3    & 92.30 $\pm$ 0.0      \\
                     & \textbf{Random} & 47.97 $\pm$ 0.5     & 46.25 $\pm$ 5.6         & 70.98 $\pm$ 0.1       & 80.64 $\pm$ 0.3   & 75.23 $\pm$ 0.1   \\ 
& \textbf{BEKM} & 50.68 $\pm$ 2.0     & 46.85 $\pm$ 4.9         & 69.64 $\pm$ 0.1       & 80.30 $\pm$ 0.2   & 74.85 $\pm$ 0.1    \\ 
& \textbf{BLPA} & 43.79 $\pm$ 2.2     & 40.24 $\pm$ 8.3         & 63.42 $\pm$ 5.7       & 85.10 $\pm$ 0.3   & 78.93 $\pm$ 0.5    \\
& \textbf{IDEA} & \textbf{72.08 $\pm$ 1.2}     & \textbf{61.56 $\pm$ 1.2}         & \textbf{73.11 $\pm$ 0.0}       & \textbf{86.13 $\pm$ 0.4}   & \textbf{91.93 $\pm$ 0.1}    \\ 
                     \midrule
\multirow{5}{*}{\textbf{SGC}} & \textbf{Re-Training} & 76.14 $\pm$ 0.6     & 65.77 $\pm$ 0.0         & 75.90 $\pm$ 0.0       & 87.10 $\pm$ 0.1    & 92.01 $\pm$ 0.0  \\
                     & \textbf{Random} & 46.00 $\pm$ 0.7     & 45.25 $\pm$ 3.0         & 69.03 $\pm$ 0.1       & 81.30 $\pm$ 0.3   & 80.81 $\pm$ 0.1   \\ 
& \textbf{BEKM} & 48.83 $\pm$ 1.8     & 46.45 $\pm$ 0.4         & 69.76 $\pm$ 0.1       & 80.08 $\pm$ 0.2   & 74.87 $\pm$ 0.2    \\ 
& \textbf{BLPA} & 63.59 $\pm$ 1.4     & 39.44 $\pm$ 2.8         & 62.98 $\pm$ 4.6       & 86.95 $\pm$ 0.1   & 87.38 $\pm$ 0.1    \\ 
& \textbf{IDEA} & \textbf{72.94 $\pm$ 1.9}     & \textbf{63.16 $\pm$ 1.0}         & \textbf{73.63 $\pm$ 0.8}       & \textbf{84.68 $\pm$ 0.3}   & \textbf{91.21 $\pm$ 0.1}  \\  
                     \midrule
\multirow{5}{*}{\textbf{GIN}} & \textbf{Re-Training} & 82.90 $\pm$ 0.6     & 74.27 $\pm$ 0.5         & 85.31 $\pm$ 0.6       & 90.28 $\pm$ 0.2    & 95.57 $\pm$ 0.2   \\
                     & \textbf{Random} & 69.25 $\pm$ 6.3     & 51.85 $\pm$ 2.7         & 83.64 $\pm$ 1.2       & 89.17 $\pm$ 0.1   & 91.74 $\pm$ 0.5   \\ 
& \textbf{BEKM} & 74.05 $\pm$ 3.5     & 65.17 $\pm$ 2.8         & 84.35 $\pm$ 0.3       & 89.39 $\pm$ 0.5   & 92.30 $\pm$ 0.3    \\ 
& \textbf{BLPA} & 62.48 $\pm$ 2.9     & 55.06 $\pm$ 7.2         & 82.25 $\pm$ 1.6       & 62.29 $\pm$ 0.7   & 71.66 $\pm$ 1.4    \\
& \textbf{IDEA} & \textbf{72.57 $\pm$ 2.8}     & \textbf{66.37 $\pm$ 4.6}         & \textbf{82.33 $\pm$ 0.2}     & \textbf{88.48 $\pm$ 0.6}    & \textbf{94.63 $\pm$ 0.1}      \\ 
                     \midrule
\multirow{5}{*}{\textbf{GAT}} & \textbf{Re-Training} & 83.76 $\pm$ 0.3     & 75.88 $\pm$ 0.1         & 85.02 $\pm$ 0.1       & 92.24 $\pm$ 0.1    & 95.28 $\pm$ 0.1         \\
                     & \textbf{Random} & 58.18 $\pm$ 2.0     & 55.43 $\pm$ 4.3         & 68.20 $\pm$ 6.9       & 80.75 $\pm$ 0.1   & 78.26 $\pm$ 0.1   \\ 
& \textbf{BEKM} & 64.20 $\pm$ 1.5     & 57.35 $\pm$ 2.8         & 71.67 $\pm$ 0.2       & 80.37 $\pm$ 0.3   & 77.47 $\pm$ 0.2    \\ 
& \textbf{BLPA} & 60.88 $\pm$ 1.0     & 58.26 $\pm$ 2.6         & 67.34 $\pm$ 3.4       & 85.22 $\pm$ 0.2   & 86.12 $\pm$ 0.2    \\ 
& \textbf{IDEA} & \textbf{84.38 $\pm$ 0.6}     & \textbf{75.78 $\pm$ 0.9}         & \textbf{84.92 $\pm$ 0.2}       & \textbf{92.20 $\pm$ 0.2}    & \textbf{95.41 $\pm$ 0.0}     \\ 
                     \bottomrule       
\end{tabular}
\vskip -2ex
\end{table*}

\subsection{Evaluation of Bound Tightness}
\label{bound_tightness_section}

To answer \textbf{RQ1}, we first evaluate how tight the derived bound between $\tilde{\bm{\theta}}^*$ and $\bar{\bm{\theta}}^*$ can be across different GNNs, graph datasets, and unlearning ratios. We also compare the bound derived based on IDEA and other bounds in existing works. To the best of our knowledge, CEU~\cite{wu2023certified} is the only existing certified unlearning approach that provides generalizable bounds across different GNNs. In particular, CEU provides bounds over the objective function after unlearning, and we adapt such bounds over the objective function to $\ell_2$ distance bounds between $\tilde{\bm{\theta}}^*$ and $\bar{\bm{\theta}}^*$ based on the common assumption of the objective function being Lipschitz continuous~\cite{wu2023certified}.
We compare the bounds and the $\ell_2$ distances below. 
\textit{(1) CEU Worst Bound.} We compute the theoretical worst bound derived based on CEU as a baseline of the $\ell_2$ distance bound between $\tilde{\bm{\theta}}^*$ and $\bar{\bm{\theta}}^*$.
\textit{(2) CEU Data-Dependent Bound.} We compute the data-dependent bound derived based on CEU as a baseline of the $\ell_2$ distance bound between $\tilde{\bm{\theta}}^*$ and $\bar{\bm{\theta}}^*$. A data-dependent bound is tighter than the Worst Bound.
\textit{(3) IDEA Bound.} We compute the bound given by Proposition~\ref{final_triangle} as the bound for the $\ell_2$ distance between $\tilde{\bm{\theta}}^*$ and $\bar{\bm{\theta}}^*$. 
\textit{(4) Actual Values.} We compare the bounds above with the actual $\ell_2$ distance between $\tilde{\bm{\theta}}^*$ and $\bar{\bm{\theta}}^*$.
Note that we focus on edge unlearning tasks to analyze the tightness of the derived bounds, since this is the only unlearning task CEU supports. We use \textit{Unlearn Ratio} to refer to the ratio of edges to be unlearned from the GNN.

We present the bounds and the actual value of the $\ell_2$ distance between $\tilde{\bm{\theta}}^*$ and $\bar{\bm{\theta}}^*$ over a wide range of unlearn ratios (from 1\% to 10\%), which covers common values, in Figure~\ref{gcn_bounds}. We also have similar observations in other cases (see Appendix). We summarize the observations below.
(1) From the perspective of the general tendency, we observe that larger unlearn ratios usually lead to larger values in both the derived bounds and the actual $\ell_2$ distance between $\tilde{\bm{\theta}}^*$ and $\bar{\bm{\theta}}^*$. This reveals that a larger unlearn ratio tends to make the approximation of $\tilde{\bm{\theta}}^*$ (with the calculated $\bar{\bm{\theta}}^*$) more difficult, which is in alignment with existing works~\cite{wu2023certified}.
(2) From the perspective of the bound tightness, we found that IDEA is able to give tighter bounds in all cases compared with the bounds given by CEU, especially in cases with larger unlearn ratios. This reveals that the approximation of $\tilde{\bm{\theta}}^*$ given by IDEA can better characterize the difference between $\tilde{\bm{\theta}}^*$ and $\bm{\theta}^*$ compared with CEU. 


\subsection{Evaluation of Unlearning Efficiency}
\label{efficiency_section}

To answer \textbf{RQ2}, we then evaluate the efficiency of IDEA in performing unlearning. Specifically, we adopt the common node unlearning task as an example, and we measure the running time of unlearning in seconds. We note that CGU only supports performing unlearning on SGC, and thus we adopt SGC as the backbone GNN for IDEA and all other baselines for a fair comparison. We use \textit{Unlearn Ratio} to refer to the ratio of training nodes to be unlearned from the GNN.
Here we present a comparison between IDEA and baselines on Cora dataset in Figure~\ref{unlearning_time}. 
We also have similar observations on other GNNs and datasets (see Appendix). We summarize the observations below. (1) From the perspective of the general tendency, we observe that the running time of re-training does not change across different unlearning ratios. This is because the number of optimization epochs dominates the running time of re-training, while the total epoch number does not change no matter how many training nodes are removed. However, the efficiency of all other baselines is sensitive to the unlearning ratio, and this is because their running time is closely dependent on the total number of nodes to be unlearned. Finally, we found that the running time of IDEA is not sensitive to the unlearn ratio. This is because the number of nodes to be unlearned will only marginally influence the computational costs associated with Theorem~\ref{influence}. The stable running time across different numbers of nodes to be unlearned serves as a key superiority of IDEA over other baselines.
(2) From the perspective of time comparison, we found that IDEA achieves significant superiority over all other baselines across the wide range of unlearning ratios, especially on relatively large ones (e.g., 10\%). Such an observation indicates that IDEA is able to perform unlearning with satisfying efficiency, which further reveals its practical significance in real-world applications.

\begin{figure}[t]
    \centering
\includegraphics[width=.48\textwidth]{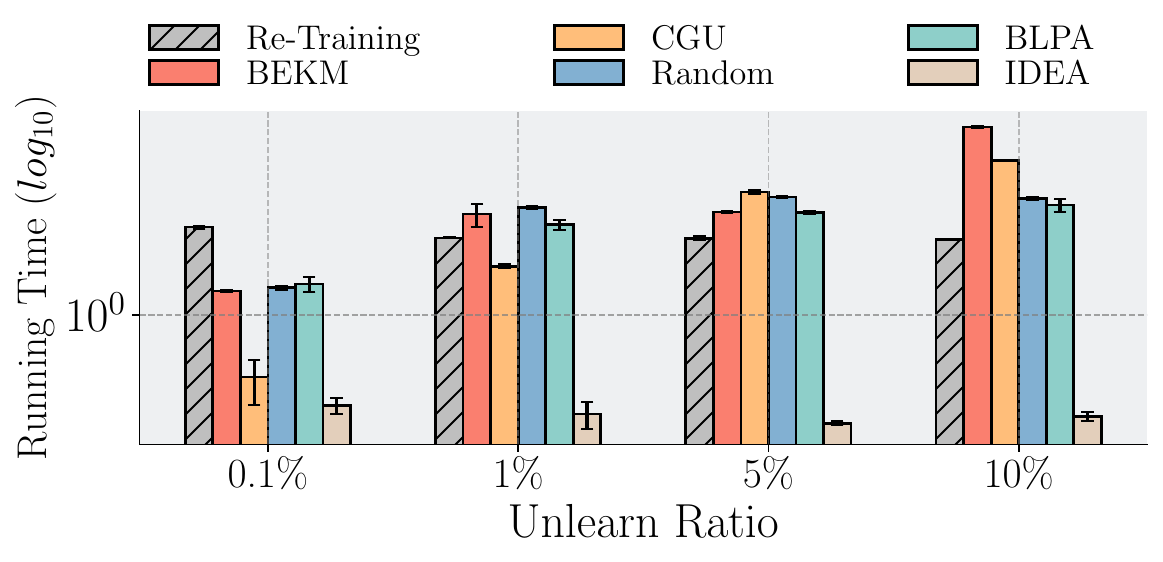}
    \vspace{-4mm}
    \caption{Efficiency comparison between IDEA and other baselines including retraining. Running time is measured with seconds and presented in log scale.} 
    \vspace{-3mm}
  \label{unlearning_time}
\end{figure}

\subsection{Evaluation of Model Utility}
\label{utility_section}

To answer \textbf{RQ3}, we now compare model utility after performing unlearning with IDEA and other baselines. We note that GraphEraser is the only baseline that supports flexible generalization across different GNN backbones. Therefore, we adopt the three variants of GraphEraser, i.e., Random, BEKM, and BLPA, as the corresponding baselines for comparison. 
We adopt the most common task of node unlearning, and we adopt the F1 score (of node classification) to measure the model utility after re-training/unlearning.
We present comprehensive empirical results (including four different GNN backbones and all five real-world datasets) in Table~\ref{unlearning_utility}. In addition to the baselines, we also report the performance of re-training, i.e., the F1 score given by a re-trained model with the unlearned nodes being removed from the training graph, for comparison.
We summarize the observations below, and similar observations are also found in different settings (see Appendix). (1) From the perspective of the general tendency, we observe that unlearning approaches are usually associated with worse utility performance compared with re-training. Such a sacrifice is usually considered acceptable, since these unlearning approaches can bring significant improvement in efficiency compared with re-training.
(2) From the perspective of model utility, we found that IDEA achieves competitive utility compared with other baselines. Specifically, compared with re-training, IDEA only sacrifices limited utility performance in most cases, and even shows better performance in certain cases. Furthermore, compared with other alternatives, IDEA also shows consistent superiority in most cases. 




\subsection{Evaluation of Unlearning Effectiveness}

\label{effectiveness_section}

To answer \textbf{RQ4}, we compare the unlearning effectiveness of IDEA and other baselines.  Specifically, we utilize the state-of-the-art attack methods MIA-Graph and StealLink to evaluate the unlearning effectiveness of node and edge unlearning tasks, respectively. To also have CGU as a baseline, we adopt SGC as the backbone GNN to ensure a fair comparison.
We present the attack successful rates after node and edge unlearning in Table~\ref{unlearning_effectiveness1}. All attacks are performed over those unlearned nodes/edges, and thus a lower AUC score represents better unlearning performance.
In terms of node attributes unlearning, we note that to the best of our knowledge, no existing membership inference attack method supports the associated attack. Here we use the average loss value as an unlearning performance indicator.
Specifically, we perform partial attribute unlearning under different ratios (20\%, 50\%, 80\%) of unlearn attribute dimensions to the total attribute dimensions. Note that partial attribute unlearning aims to twist the GNN model such that the GNN model behaves as if it were trained on those nodes with the unlearn attribute values being set to non-informative numbers (as in Section~\ref{request_instantiations}). Here we follow a common choice~\cite{chien2022certified} to set such a number as zero.
Accordingly, we evaluate the performance with the average loss values regarding the nodes with the unlearn attributes being set to zeros, and a lower loss value indicates better unlearning effectiveness.
We present the results in Table~\ref{unlearning_effectiveness2}. Note that CGU only supports full attribute unlearning, while the three variants of GraphEraser only support node/edge unlearning. Therefore, we perform attribute unlearning and node unlearning for CGU and GraphEraser, respectively.
Based on the settings above, we have the observations below, and consistent observations are also found under different settings (see Appendix). 
(1) From the perspective of node and edge unlearning, we observe that the attack AUC scores over IDEA are among the lowest in both unlearning tasks. Noticing that the AUC scores given by IDEA are only marginally above 50\%, the unlearned node/edge information has been almost completely removed from the trained GNNs.
(2) From the perspective of attribute unlearning, IDEA exhibits the lowest average loss values in all (attribute) unlearn ratios. This indicates the superior attribute unlearning performance.

\begin{table}[t]
\setlength{\tabcolsep}{6.65pt}
\renewcommand{\arraystretch}{0.99}
\centering
\vspace{1mm}
\caption{Attack AUC scores after node and edge unlearning on Cora. The results given by IDEA are marked in bold.}
\label{unlearning_effectiveness1}
\begin{tabular}{lcc}
\toprule
       & \textbf{Node Unlearning ($\downarrow$)} & \textbf{Edge Unlearning ($\downarrow$)}  \\
       \midrule
\textbf{Random} & 50.38 $\pm$ 0.5     & 55.64 $\pm$ 2.8        \\
\textbf{BEKM}   &  50.35 $\pm$ 1.2  & 51.81 $\pm$ 0.3               \\
\textbf{BLPA}   & 50.30 $\pm$ 0.4     & 50.84 $\pm$ 3.4              \\
\textbf{CGU}    & 54.67 $\pm$ 2.9     & 66.52 $\pm$ 0.6                \\
\textbf{IDEA}   &  \textbf{50.86 $\pm$ 1.8}    &  \textbf{50.11 $\pm$ 0.9}            \\
\bottomrule
\end{tabular}
\vskip -0.7ex
\end{table}

\begin{table}[t]
\setlength{\tabcolsep}{9.9pt}
\renewcommand{\arraystretch}{0.99}
\centering
\caption{Average loss values on Cora regarding the nodes with the unlearn attributes being set to zeros. Ratio of unlearn node attribute dimensions to all attribute dimensions varies across 20\%, 50\%, and 80\%. Lower values represent better performance, and results from IDEA are marked in bold.}
\label{unlearning_effectiveness2}
\begin{tabular}{lccc}
\toprule
       & \textbf{20\% ($\downarrow$)} & \textbf{50\% ($\downarrow$)} & \textbf{80\% ($\downarrow$)} \\
       \midrule
\textbf{Random} & 1.32 $\pm$ 0.06     & 1.38 $\pm$ 0.06     & 1.35 $\pm$ 0.09          \\
\textbf{BEKM}   &  1.41  $\pm$ 0.16   & 1.47  $\pm$ 0.14    & 1.39  $\pm$ 0.12               \\
\textbf{BLPA}   &  1.47 $\pm$ 0.11   &  1.69 $\pm$ 0.37   &  1.50  $\pm$ 0.05              \\
\textbf{CGU}    &  1.62 $\pm$ 0.02   & 1.73 $\pm$ 0.04   & 1.78    $\pm$ 0.06            \\
\textbf{IDEA}    & \textbf{1.29 $\pm$ 0.01}    & \textbf{1.31 $\pm$ 0.01}    & \textbf{1.33 $\pm$ 0.01}                \\
\bottomrule
\end{tabular}
\vskip -3ex
\end{table}

\section{Related Work}

\noindent \textbf{Certified Machine Unlearning.}
The general desiderata of machine unlearning is to remove the influence of certain training data on the model parameters, such that the model can behave as if it never saw such data~\cite{xu2023machine,nguyen2022survey,bourtoule2021machine,yan2022arcane,wang2024machine,grimes2024gone}. Re-training the model without making the unlearning data visible is an ideal way to achieve such a goal, while it is usually infeasible in practice due to various reasons such as prohibitively high computational costs~\cite{zhang2023review,xu2024machine,qu2023learn,schelter2023forget}. A popular way to approach the goal of unlearning is to directly approximate the re-trained model parameters, a.k.a., \textit{approximate unlearning}~\cite{xu2023machine,thudi2022unrolling}. Certified machine unlearning is under the umbrella of approximate unlearning~\cite{liu2024threats,zhang2023survey}, and it has stood out due to the capability of providing theoretical guarantee on the unlearning effectiveness. A commonly used criterion of certified unlearning is \textit{$(\varepsilon-\delta)$ certified unlearning}~\cite{guo2019certified,sekhari2021remember,liu2024certified,dai2022comprehensive}, which utilizes two parameters $\varepsilon$ and $\delta$ to describe the proximity between the re-trained model parameter distribution and approximated model parameter distribution in the model space. In recent years, various techniques have been proposed to achieve certified unlearning~\cite{zhang2022prompt,warnecke2021machine,mahadevan2021certifiable}. However, they overwhelmingly focus on independent, identically distributed (i.i.d.) data and fail to consider the dependency between data points. Therefore, they cannot be directly adopted to perform unlearning over GNNs~\cite{wu2023gif,wu2023certified}. Different from the works mentioned above, our paper proposes a certified unlearning approach for GNNs, necessitating the modeling of dependencies between instances in graphs (e.g., nodes and edges).

\noindent \textbf{Machine Unlearning for Graph Neural Networks.} Over the years, GNNs have been increasingly deployed in a plethora of applications~\cite{zhou2020graph,wu2022graph,neo2024towards,jin2023predicting,ma2022assessing,shi2023gigamae,dong2023reliant,zhang2023adversarial,dong2023graph}. Similar to other machine learning models, these GNN models also face the risk of privacy leakage~\cite{qiu2022privacy,shan2021towards,ju2024survey,sajadmanesh2023privacy,wang2023fair,xu2023mdp}, where the private information is usually considered to be encoded in the training data~\cite{said2023survey}. Such a threat has prompted the emerging of unlearning approaches for GNNs~\cite{chen2022graph,pan2023unlearning,cheng2023gnndelete,wu2023gif}. However, these works are only able to achieve unlearning for GNNs empirically, failing to provide any theoretical guarantee on the effectiveness. To further strengthen the power of unlearning for GNNs and enhance the confidence of model owners before model deployment, a few recent works have initiated explorations on certified unlearning for GNNs. Wu et al.~\cite{wu2023certified} propose CEU to unlearn edges that are visible to GNNs during training, while edge unlearning is the only type of request it is able to handle. Chien et al.~\cite{chien2022certified} proposed a different certified unlearning approach for GNNs to also handle node and attribute unlearning requests, while such an approach is only applicable to a specially simplified GNN model. Meanwhile, these approaches can only handle limited types of unlearning requests, which further jeopardizes their flexibility in real-world applications. Different from them, our paper proposes a flexible unlearning framework that can handle different types of unlearning requests. On top of this framework, an effectiveness certification is further proposed without relying on any specific GNN structure or objective function.

\section{Conclusion}

In this paper, we propose IDEA, a flexible framework of certified unlearning for GNNs. Specifically, we first formulate and study a novel problem of flexible and certified unlearning for GNNs, which aims to flexibly handle different unlearning requests with theoretical guarantee. To tackle this problem, we develop IDEA by analyzing the objective difference before and after certain information is removed from the graph. We further present theoretical guarantee as the certification for unlearning effectiveness. Extensive experiments on real-world datasets demonstrate the superiority of IDEA in multiple key perspectives. Meanwhile, two future directions are worth further investigation. First, we focus on the common node classification task in this paper, and we will extend the proposed framework to other tasks, such as graph classification. Second, considering that GNNs may be trained in a decentralized manner, it is critical to study GNN unlearning under a distributed setting. 

\section{Acknowledgements}

This work is supported in part by the National Science Foundation under IIS-2006844, IIS-2144209, IIS-2223769, IIS-1900990, IIS-2239257, IIS-2310262, CNS-2154962, and BCS-2228534; the Commonwealth Cyber Initiative Awards under VV-1Q23-007, HV-2Q23-003, and VV-1Q24-011; the JP Morgan Chase Faculty Research Award; the Cisco Faculty Research Award; and Snap gift funding.

\bibliographystyle{ACM-Reference-Format}
\bibliography{references}


\clearpage
\appendix


\section{Reproducibility}
\label{appendix_implementation}

In this section, we introduce the details of the experiments in this paper for the purpose of reproducibility. At the same time, we have uploaded all necessary code to our GitHub repository to reproduce the results presented in this paper: \textcolor{black}{\url{https://github.com/yushundong/IDEA}}. All major experiments are encapsulated as shell scripts, which can be conveniently executed. We introduce details in the subsections below.

\subsection{Real-World Datasets.}

Here we briefly introduce the five real-world graph datasets we used in this paper, and all these datasets are commonly used datasets in node classification tasks. We present their statistics in Table~\ref{datasets_stats}.

\subsection{Experimental Settings.}

For all datasets, we follow a commonly used split~\cite{wu2023gif} of 90\% and 10\% for training and test node set, respectively. In the task of node classification, only the node labels in the training set are visible for all models during the training process. We adopt a learning rate of 0.01 for most models in our experiments, including the backbone GNNs in the proposed framework IDEA, together with commonly used number of epochs, e.g., 300. The value of the standard deviation adopted by IDEA falls into the range of (1e-3, 1e0), and the values are consistent in different experiments for each unlearning task.

\subsection{Implementation of IDEA.}

IDEA is implemented based on PyTorch~\cite{paszke2017automatic} and optimized through Adam optimizer~\cite{kingma2014adam}. In experiments, IDEA and other unlearning baselines require assumption on the characteristics regarding the objective function. We set all these values to be commonly used values in existing works focusing on machine unlearning, and these values are set to be consistent across all methods for the purpose of fair comparison. Here we present the specific values we adopted. Specifically, the objective function is assumed to be 0.05 strongly-convex, and the Lipschitz constant of the objective function is set to be 0.25. The numerical bound of the training loss regarding each training nodes is set to be 3.0 around the optimal point, which can be easily satisfied in practice. In addition, the Lipschitz constant of the Hessian matrix of the objective function is assumed to be 0.25. The derivative of the objective function is set to be bounded by 1.0, and the Lipschitz constant of the first-order derivative of the objective function is also set to be 1.0.

\subsection{Implementation of Graph Neural Networks.}

For all adopted GNNs (i.e., SGC, GCN, GAT, and GIN) in our experiments, their released implementations are utilized for a fair comparison. The layer number for GCN and GIN is set as two. We adopt a dropout rate as 0.6, and the attention head number of GAT is set as eight. Meanwhile, we note that same observations can also be found under different settings.

\begin{table}[]
\setlength{\tabcolsep}{5.9pt}
\renewcommand{\arraystretch}{1.1}
\centering
\vspace{2mm}
\caption{Statistics of the adopted real-world graph datasets.}
\vspace{-3mm}
\label{datasets_stats}
\begin{tabular}{lcccc}
\toprule
         & \#\textbf{Nodes} & \#\textbf{Edges} & \#\textbf{Attributes} & \#\textbf{Classes} \\
         \midrule
\textbf{Cora}     & 2,708        &  5,429       &   1,433           &   7       \\
\textbf{CiteSeer} & 3,327        &  4,723       &  3,703            &   6      \\
\textbf{PubMed}   & 19,717        & 88,648        & 500             &  3       \\
\textbf{CS}       & 18,333        & 163,788        & 6,805             & 15       \\
\textbf{Physics}  & 34,493        & 495,924        & 8,415             &  5       \\
\bottomrule
\end{tabular}
\end{table}

\subsection{Implementation of Baselines.}


\textbf{Re-Training.} We adopt re-training as our first baseline for performance comparison. Note that the randomness in initialization and optimization may significantly influence the performance. Hence we continue to train the optimized GNN model with the unlearning instances (e.g., nodes and edges to be unlearned) being removed, and consider this model as the re-trained model. Such a strategy is also widely used in other works (e.g., \cite{koh2017understanding}) based on influence function and their main goal remains consistent with us, i.e., to avoid the randomness induced by initialization and optimization.

\noindent \textbf{Graph Unlearning (GraphEraser).} We adopt GraphEraser as another baseline for comparison. In total, GraphEraser has three types of variants (including \texttt{Random}, \texttt{BEKM}, \texttt{BLPA}), which are all adopted in our paper for comparison. These variants associates with different types of graph partition methods. In addition, we follow their reported results to adopt \texttt{LBAggr} as the aggregator in GraphEraser to achieve its best performance in model utility. We adopt its official open-source code\footnote{https://github.com/MinChen00/Graph-Unlearning} for experiments.

\noindent \textbf{Certified Graph Unlearning (CGU).} We adopt CGU as another baseline for comparison. We note that CGU only supports SGC as the corresponding GNN backbone model. We adopt its official open-source code\footnote{https://github.com/thupchnsky/sgc\_unlearn} for experiments.

\noindent \textbf{Certified Edge Unlearning (CEU).} We also adopt CEU as a baseline for comparison. We note that CEU only supports edge unlearning as the corresponding unlearning task. We adopt its official open-source code\footnote{https://github.com/kunwu522/certified\_edge\_unlearning} for experiments.

\subsection{Implementation of Threat Models.}


\noindent \textbf{MIA-Graph.} We adopt MIA-Graph as the node attack method. We note that since MIA-Graph has several variants that might influence attack efficacy, we select the attack strategy with the best empirical performance, which entails training a shadow model using the original dataset's ground truth labels. Then we employ the output class posterior probabilities of the shadow model to train an attack model, which is later deployed to execute attacks on the target model with unlearned instances. To compute the successful rate, we select all the unlearned instances as positive samples, and an equal number of instances at random from the target model's test set as negative samples. We adopt the official code\footnote{https://github.com/iyempissy/rebMIGraph} for experiments.

\noindent \textbf{StealLink.} We adopt StealLink as the edge attack method. Note that StealLink integrates multiple attack strategies each necessitating different background knowledge. Here we choose attack-3 from the original paper, wherein the adversary has access to a partial graph of the target datasets, due to its high successful rate in attacks. Here we select a partial graph size equivalent to $50\%$ of all edges to train the attack model, where the output class posterior probabilities are utilized as features. To compute the successful rate, we select all the unlearned edges as positive samples, and an equal number of instances at random from the edges that do not exist in the original dataset. We adopt official open-source code\footnote{https://github.com/xinleihe/link\_stealing\_attack} for experiments.

\subsection{Packages Required for Implementations.}
We perform the experiments on a server with multiple Nvidia A6000 GPUs. Below we list the key packages and their associated versions in our implementation.
\begin{itemize}[topsep=4pt]
    \item Python == 3.8.8
    \item torch == 1.10.1+cu111
    \item torch-cluster == 1.6.0
    \item torch-geometric == 2.2.0
    \item torch-scatter == 2.0.9
    \item torch-sparse == 0.6.13
    \item cuda == 11.1
    \item numpy == 1.20.1
    \item tensorboard == 1.13.1
    \item networkx == 2.5
    \item scikit-learn==0.24.2
    \item pandas==1.2.4
    \item scipy==1.6.2
\end{itemize}

\section{Proofs}
\label{appendix_proofs}

\begin{appendix_proposition}
\textbf{Localized Equivalence of Training Nodes.} Given $\Delta \mathcal{G} = \{\Delta \mathcal{V}, \Delta \mathcal{E}, \Delta \mathcal{X}\}$ to be unlearned ($v_i \notin \Delta \mathcal{V}$) and an objective $\mathscr{L}$ computed over $f_{\bm{\theta}}$, $\mathscr{L}\left(\bm{\theta}, v_i, \mathcal{G}\right) = \mathscr{L}\left(\bm{\theta}, v_i, \mathcal{G} \ominus \Delta\mathcal{G}\right)$ holds $\forall v_i \notin \phi_k(v_j) \cup \{v_j\}, v_j \in \Delta \mathcal{V}  \cup \gamma_e(\Delta \mathcal{E}) \cup \gamma_x(\Delta \mathcal{X})$. Here $\gamma_e$ and $\gamma_x$ return the set of nodes that directly connect to the edges in $\mathcal{E}$ and that have associated attribute(s) in $\mathcal{X}$, respectively.
\end{appendix_proposition}

\begin{proof}
For most GNNs based on message passing, the embedding of a node is only determined by its $k$-hop neighbors~\cite{hamilton2017inductive}. For each layer in the GNN, each node can be seen as aggregating the embeddings of its one-hop neighbors.
For any $v_j\in\Delta\mathcal{V}$, $v_i\notin\phi_k(v_j)$ and we have $v_j\notin\phi_k(v_i)$.
Consequently, we have $\mathscr{L}\left(\bm{\theta}, v_i, \mathcal{G}\right) = \mathscr{L}\left(\bm{\theta}, v_i, \mathcal{G} \ominus \Delta\mathcal{G}\right)$.
\end{proof}

\begin{appendix_lemma}
\textbf{Optimal Equivalence.} The optimal solution to Equation (\ref{changed_obj}) (denoted as $\bm{\theta}_{\Delta \mathcal{G}, \xi}^{*}$) equals to the optimal solution to Equation (\ref{new_obj}) (denoted as $\tilde{\bm{\theta}}^{*}$) when $\xi = \frac{1}{m}$.
\end{appendix_lemma}

\begin{proof}
Incorporate \Cref{l_add} and \Cref{l_sub} into \Cref{changed_obj}, we come to
\begin{equation}
\begin{aligned}
\bm{\theta}_{\Delta \mathcal{G},\xi}^{*}&=\arg\min_{\bm{\theta}}\frac{1}{m}\sum_{v_i\in\mathcal{V}_{\text{trn}}}\mathscr{L}\left(\bm{\theta},v_i,\mathcal{G}\right)+\xi\left(\mathscr{L}_{\text{add}}-\mathscr{L}_{\text{sub}}\right) \\
&=\arg\min_{\bm{\theta}}\frac{1}{m}\sum_{v_i\in\mathcal{V}_{\text{trn}}}\mathscr{L}\left(\bm{\theta},v_i,\mathcal{G}\right) \\
&+\xi\left(\sum_{v_i\in \mathcal{V}_{\text{trn}}\backslash\Delta\mathcal{V}}\mathscr{L}\left(\bm{\theta},v_i,\mathcal{G}\ominus\Delta\mathcal{G}\right)-\sum_{v_i\in\mathcal{V}_{\text{trn}}}\mathscr{L}\left(\bm{\theta},v_i,\mathcal{G}\right)\right).
\end{aligned}
\end{equation}
When $\xi=\frac{1}{m}$, we consequently have
\begin{equation}
\bm{\theta}_{\Delta \mathcal{G},\xi}^{*}=\arg\min_{\bm{\theta}}\frac{1}{m}\sum_{v_i\in \mathcal{V}_{\text{trn}}\backslash\Delta\mathcal{V}}\mathscr{L}\left(\bm{\theta},v_i,\mathcal{G}\ominus\Delta\mathcal{G}\right)=\tilde{\bm{\theta}}^*,
\end{equation}
which finishes our proof.
\end{proof}

\begin{appendix_theorem}
\textbf{Approximation with Infinitesimal Residual.} Given a graph data $\mathcal{G}$, $\Delta \mathcal{G} = \{\Delta \mathcal{V}, \Delta \mathcal{E}, \Delta \mathcal{X}\}$ to be unlearned, and an objective $\mathscr{L}$ computed over an $f_{\bm{\theta}^*}$, using $\bm{\theta}^* + \frac{1}{m} \Delta \bar{\bm{\theta}}^*$ as an approximation of $\tilde{\bm{\theta}}^*$ only brings a first-order infinitesimal residual w.r.t. $\|\bm{\theta}^* - \tilde{\bm{\theta}}^*\|_2$, where $\Delta \bar{\bm{\theta}}^* = -\bm{H}^{-1}_{\bm{\theta}^*} \left( \nabla_{\bm{\theta}} \mathscr{L}_{\text{add}} - \nabla_{\bm{\theta}} \mathscr{L}_{\text{sub}}\right)$, and $\bm{H}_{\bm{\theta}^*} \coloneqq \nabla_{\bm{\theta}}^{2} \frac{1}{m} \sum_{v_i \in \mathcal{V}_{\text{trn}}} \mathscr{L}\left(\bm{\theta}, v_i, \mathcal{G}\right)$. 
\end{appendix_theorem}

\begin{proof}
For simplicity, we define the function $\Phi(\cdot)$ as $\Phi(\bm{\theta})=\frac{1}{m} \sum_{v_i \in \mathcal{V}_{\text{trn}}} \mathscr{L}\left(\bm{\theta}, v_i, \mathcal{G}\right) + \xi \left( \mathscr{L}_{\text{add}} - \mathscr{L}_{\text{sub}}\right)$.
We then conduct the Taylor expansion of $\nabla_{\bm{\theta}}\Phi(\bm{\theta})$ at $\bm{\theta}=\bm{\theta}^*$ as
\begin{equation}\label{eq:taylor_expansion}
\nabla_{\bm{\theta}}\Phi(\bm{\theta})=\nabla_{\bm{\theta}}\Phi(\bm{\theta}^*)+\nabla^2_{\bm{\theta}}\Phi(\bm{\theta}^*)(\bm{\theta}-\bm{\theta}^*)+o(\|\bm{\theta}-\bm{\theta}^*\|).
\end{equation}
Then we let $\bm{\theta}=\bm{\theta}^*_{\Delta\mathcal{G},\xi}$ and have
\begin{equation}
\nabla_{\bm{\theta}}\Phi(\bm{\theta}^*)+\nabla^2_{\bm{\theta}}\Phi(\bm{\theta}^*)(\bm{\theta}^*_{\Delta\mathcal{G},\xi}-\bm{\theta}^*)=o(\|\bm{\theta}^*_{\Delta\mathcal{G},\xi}-\bm{\theta}^*\|),
\end{equation}
based on the fact that $\nabla_{\bm{\theta}}\Phi(\bm{\theta}^*_{\Delta\mathcal{G},\xi})=0$.
Consequently, we have
\begin{equation}\label{eq:influence}
\bm{\theta}^*_{\Delta\mathcal{G},\xi}-\bm{\theta}^*=-\nabla^{-2}_{\bm{\theta}}\Phi(\bm{\theta}^*)\left(\nabla_{\bm{\theta}}\Phi(\bm{\theta}^*)+o(\|\bm{\theta}^*_{\Delta\mathcal{G},\xi}-\bm{\theta}^*\|)\right).
\end{equation}
We first take a look at the term $\nabla^{-2}_{\bm{\theta}}\Phi(\bm{\theta}^*)$. In particular, we have
\begin{equation}\label{eq:second_order}
\begin{aligned}
\nabla^{-2}_{\bm{\theta}}\Phi(\bm{\theta}^*)&=\left(\frac{1}{m}\sum_{v_i\in\mathcal{V}_{\text{trn}}}\nabla^2_{\bm{\theta}}\mathscr{L}\left(\bm{\theta}^*,v_i,\mathcal{G}\right)+\xi\nabla^2_{\bm{\theta}}\left(\mathscr{L}_{\text{add}}-\mathscr{L}_{\text{sub}}\right)\right)^{-1} \\
&=\left(\frac{1}{m}\sum_{v_i\in\mathcal{V}_{\text{trn}}}\nabla^2_{\bm{\theta}}\mathscr{L}\left(\bm{\theta}^*,v_i,\mathcal{G}\right)\right)^{-1}+o(\xi),
\end{aligned}
\end{equation}
where the second equality holds according to Taylor's theorem. 
We then take a look at the term $\nabla_{\bm{\theta}}\Phi(\bm{\theta}^*)$ and have
\begin{equation}\label{eq:first_order}
\begin{aligned}
\nabla_{\bm{\theta}}\Phi(\bm{\theta}^*)&=\frac{1}{m}\sum_{v_i\in\mathcal{V}_{\text{trn}}}\nabla_{\bm{\theta}}\mathscr{L}\left(\bm{\theta}^*,v_i,\mathcal{G}\right)+\xi\cdot\nabla_{\bm{\theta}}\left(\mathscr{L}_{\text{add}}-\mathscr{L}_{\text{sub}}\right) \\
&=\xi\cdot\nabla_{\bm{\theta}}\left(\mathscr{L}_{\text{add}}-\mathscr{L}_{\text{sub}}\right),
\end{aligned}
\end{equation}
where the second equality holds based on the definition of $\bm{\theta}^*$.
Given that $\xi\rightarrow0$ (and $\bm{\theta}^*_{\Delta\mathcal{G},\xi}\rightarrow\bm{\theta}^*$ consequently), we then incorporate \Cref{eq:second_order} and \Cref{eq:first_order} into \Cref{eq:influence} and have
\begin{equation}
\begin{aligned}
\bm{\theta}^*_{\Delta\mathcal{G},\xi}-\bm{\theta}^*&=-\xi\bm{H}_{\bm{\theta}^*}^{-1}\nabla_{\bm{\theta}}\left(\mathscr{L}_{\text{add}}-\mathscr{L}_{\text{sub}}\right)+o(\xi^2)+o(\|\bm{\theta}^*_{\Delta\mathcal{G},\xi}-\bm{\theta}^*\|),
\end{aligned}
\end{equation}
based on the fact that $\nabla_{\bm{\theta}}\Phi(\bm{\theta}^*)=o(\xi)\cdot\bm{1}$ and neglecting the intersection term $o(\xi\|\bm{\theta}^*_{\Delta\mathcal{G},\xi}-\bm{\theta}^*\|)$.
Based on \Cref{new_equivalence}, we have $\bm{\theta}^*_{\Delta\mathcal{G},\xi}=\tilde{\bm{\theta}}^*$ when $\xi=\frac{1}{m}$.
Finally, let $\xi=\frac{1}{m}$ and omit the higher-order infinitesimal terms, and then we have
\begin{equation}
\tilde{\bm{\theta}}^*-\bm{\theta}^*=\frac{1}{m}\Delta\bar{\bm{\theta}}^*+o(\|\bm{\theta}^*-\tilde{\bm{\theta}}^*\|),
\end{equation}
which finishes the proof.
\end{proof}

\begin{appendix_proposition} 
    \textbf{Serializability of Approximation.} Any mixture of unlearning request instantiations can be split into multiple sets of unlearning requests, where each set of unlearning requests satisfies $\forall \mathcal{V}_i \cap \mathcal{V}_j = \varnothing$ and $\forall \tilde{\mathcal{V}}_i \cap \tilde{\mathcal{V}}_j = \varnothing$ for $i, j \in \{1, 2, 3, 4\}$ when $i \neq j$. Serially performing approximation following these request sets achieves upper-bounded error.
\end{appendix_proposition}

\begin{proof}
We first prove the statement (1).
Obviously, all unlearning requests can be divided into units such as removing one node/edge or part of the features from one node.
In the worst case, we can execute one unlearning request unit each time to ensure that every two request units have no overlapping graph entities.
We then prove the statement (2).
We denote the number of unlearned nodes in each unlearning request unit as $|\Delta\mathcal{V}_{\text{unit}}|=N$ and $|\tilde{\mathcal{V}}_{\text{unit}}|=N'$ and then have
\begin{equation}\label{eq:sequantial_bound}
\begin{aligned}
\|\tilde{\bm{\theta}}^*_t-\bar{\bm{\theta}}^*_t\|_2&\leq\|\tilde{\bm{\theta}}^*_t-\bm{\theta}^*\|_2+\|\bar{\bm{\theta}}^*_t-\bm{\theta}^*\|_2 \\
&\leq\frac{tLN+\sqrt{4tm\lambda CN'+t^2L^2N^2}}{m\lambda}+\frac{1}{m}\sum_{i=1}^t\|\Delta\bar{\bm{\theta}}^*_i\|_2,
\end{aligned}
\end{equation}
where $\tilde{\bm{\theta}}^*_t$ and $\bar{\bm{\theta}}^*_t$ denote the retrained model and unlearning approximation after executing $t$ unlearning request units.
The second inequality holds according to \Cref{final_triangle}.
From the results shown in \Cref{eq:sequantial_bound}, we can obtain that the sequential approximation has a linearly increasing upper bound (w.r.t the number of unlearning request units $t$).
\end{proof}

\begin{appendix_assumption}\label{asp:loss}
The loss values of optimal points are bounded: $|\mathscr{L}(\bm{\theta}^*)|\leq C$ and $|\mathscr{L}(\tilde{\bm{\theta}}^*)|\leq C$.
\end{appendix_assumption}

\begin{appendix_assumption}\label{asp:continuity}
The loss function $\mathscr{L}$ is $L$-Lipschitz continuous.
\end{appendix_assumption}

\begin{appendix_assumption}\label{asp:convexity}
The loss function $\mathscr{L}$ is $\lambda$-strongly convex.
\end{appendix_assumption}

\begin{appendix_theorem}
\textbf{Distance Bound in Optimals.} The $\ell_2$ distance bound between $\tilde{\bm{\theta}}^*$ and $\bm{\theta}^*$ is given by 
\begin{align}
\|\tilde{\bm{\theta}}^*-\bm{\theta}^*\|_2 \leq\frac{L|\Delta\mathcal{V}|+\sqrt{4m\lambda C|\tilde{\mathcal{V}}|+L^2|\Delta\mathcal{V}|^2}}{m\lambda}.
\end{align}
Denote $\mathcal{V}_x^{\text{(F+P)}} = \mathcal{V}_x^{\text{(Full)}} \cup \mathcal{V}_x^{\text{(Partial)}}$, and $\tilde{\mathcal{V}}$ is given by
\begin{align}
\tilde{\mathcal{V}} = \mathcal{V}_1 \cup \mathcal{V}_4 \cup \{v_i: v_i \in \phi_k(v_j) \cap \mathcal{V}_{\text{trn}}, \; v_j \in \mathcal{V}_x^{\text{(F+P)}}\}.
\end{align}
\end{appendix_theorem}

\begin{proof}
For simplicity, we first denote the objective functions over original and retained graphs as $\mathcal{F}(\bm{\theta})=\frac{1}{m}\sum_{v_i\in\mathcal{V}_{\text{trn}}}\mathscr{L}(\bm{\theta},v_i,\mathcal{G})$ and $\hat{\mathcal{F}}(\bm{\theta})=\frac{1}{m-|\Delta\mathcal{V}|}\sum_{v_i\in\mathcal{V}_{\text{trn}}\backslash\Delta\mathcal{V}}\mathscr{L}(\bm{\theta},v_i,\mathcal{G}\ominus\Delta\mathcal{G})$.
In particular, we can rewrite the function $\hat{\mathcal{F}}(\bm{\theta})$ as 
\begin{equation}
\begin{aligned}
\hat{\mathcal{F}}(\bm{\theta})&=\frac{1}{m-|\Delta\mathcal{V}|}\sum_{v_i\in\mathcal{V}_{\text{trn}}\backslash\Delta\mathcal{V}}\mathscr{L}(\bm{\theta},v_i,\mathcal{G}) \\
&\quad\quad+\frac{1}{m-|\Delta\mathcal{V}|} \sum_{v_i \in \tilde{\mathcal{V}}}(\mathscr{L}(\bm{\theta},v_i,\mathcal{G}\ominus\Delta\mathcal{G}) - \mathscr{L}(\bm{\theta},v_i,\mathcal{G})), \\
\end{aligned}
\end{equation}
where $\tilde{\mathcal{V}}$ denotes the set of nodes whose representation can be affected by removing the graph entity $\mathcal{G}\ominus\Delta\mathcal{G}$ (the message of a node can only reach the $k$-hop neighbors where $k$ is the layer number of the GNN~\cite{hamilton2017inductive}). 
We then have
\begin{equation}
\begin{aligned}
&m(\mathcal{F}(\tilde{\bm{\theta}}^*)-\mathcal{F}(\bm{\theta}^*)) \\
=&\sum_{v_i\in\mathcal{V}_{\text{trn}}}\mathscr{L}(\tilde{\bm{\theta}}^*,v_i,\mathcal{G})-\sum_{v_i\in\mathcal{V}_{\text{trn}}}\mathscr{L}(\bm{\theta}^*,v_i,\mathcal{G}) \\
=&\sum_{v_i\in\mathcal{V}_{\text{trn}}\backslash\Delta\mathcal{V}}\mathscr{L}(\tilde{\bm{\theta}}^*,v_i,\mathcal{G})-\sum_{v_i\in\mathcal{V}_{\text{trn}}\backslash\Delta\mathcal{V}}\mathscr{L}(\bm{\theta}^*,v_i,\mathcal{G}) \\
&\quad\quad+\sum_{v_i\in\Delta\mathcal{V}}\mathscr{L}(\tilde{\bm{\theta}}^*,v_i,\mathcal{G})-\sum_{v_i\in\Delta\mathcal{V}}\mathscr{L}(\bm{\theta}^*,v_i,\mathcal{G}) \\
=&(m-|\Delta\mathcal{V}|)\hat{\mathcal{F}}(\tilde{\bm{\theta}}^*)+\sum_{v_i\in\tilde{\mathcal{V}}}\mathscr{L}(\tilde{\bm{\theta}}^*,v_i,\mathcal{G})-\mathscr{L}(\tilde{\bm{\theta}}^*,v_i,\mathcal{G}\ominus\Delta\mathcal{G}) \\
&-(m-|\Delta\mathcal{V}|)\hat{\mathcal{F}}(\bm{\theta}^*)-\sum_{v_i\in\tilde{\mathcal{V}}}\mathscr{L}(\bm{\theta}^*,v_i,\mathcal{G})-\mathscr{L}(\bm{\theta}^*,v_i,\mathcal{G}\ominus\Delta\mathcal{G}) \\
&\quad\quad+\sum_{v_i\in\Delta\mathcal{V}}\mathscr{L}(\tilde{\bm{\theta}}^*,v_i,\mathcal{G})-\sum_{v_i\in\Delta\mathcal{V}}\mathscr{L}(\bm{\theta}^*,v_i,\mathcal{G}). 
\end{aligned}
\end{equation}
According to the definition of $\tilde{\bm{\theta}}^*$, we assume that $\hat{\mathcal{F}}(\tilde{\bm{\theta}}^*)\approx0$ and have $\hat{\mathcal{F}}(\bm{\theta}^*)\geq0$.
Consequently, we have
\begin{equation}
\begin{aligned}
&m(\mathcal{F}(\tilde{\bm{\theta}}^*)-\mathcal{F}(\bm{\theta}^*)) \\
\leq&\sum_{v_i\in\tilde{\mathcal{V}}}\mathscr{L}(\tilde{\bm{\theta}}^*,v_i,\mathcal{G})-\mathscr{L}(\tilde{\bm{\theta}}^*,v_i,\mathcal{G}\ominus\Delta\mathcal{G}) \\
&\quad-\sum_{v_i\in\tilde{\mathcal{V}}}\mathscr{L}(\bm{\theta}^*,v_i,\mathcal{G})-\mathscr{L}(\bm{\theta}^*,v_i,\mathcal{G}\ominus\Delta\mathcal{G}) \\
&\quad\quad+\sum_{v_i\in\Delta\mathcal{V}}\mathscr{L}(\tilde{\bm{\theta}}^*,v_i,\mathcal{G})-\sum_{v_i\in\Delta\mathcal{V}}\mathscr{L}(\bm{\theta}^*,v_i,\mathcal{G}).
\end{aligned}
\end{equation}
According to \Cref{asp:loss}, we then have
\begin{equation}
\begin{aligned}
&\mathcal{F}(\tilde{\bm{\theta}}^*)-\mathcal{F}(\bm{\theta}^*) \\
\leq&\frac{1}{m}\left(\sum_{v_i\in\tilde{\mathcal{V}}}2C-\sum_{v_i\in\tilde{\mathcal{V}}}-2C+\sum_{v_i\in\Delta\mathcal{V}}\mathscr{L}(\tilde{\bm{\theta}}^*,v_i,\mathcal{G})-\mathscr{L}(\bm{\theta}^*,v_i,\mathcal{G})\right) \\
\leq&\frac{4}{m}|\tilde{\mathcal{V}}|C+\frac{1}{m}\sum_{v_i\in\Delta\mathcal{V}}\mathscr{L}(\tilde{\bm{\theta}}^*,v_i,\mathcal{G})-\mathscr{L}(\bm{\theta}^*,v_i,\mathcal{G}) \\
\leq&\frac{4}{m}|\tilde{\mathcal{V}}|C+\frac{1}{m}\sum_{v_i\in\Delta\mathcal{V}}|\mathscr{L}(\tilde{\bm{\theta}}^*,v_i,\mathcal{G})-\mathscr{L}(\bm{\theta}^*,v_i,\mathcal{G})|.
\end{aligned}
\end{equation}
Based on \Cref{asp:continuity}, we then come to
\begin{equation}
\mathcal{F}(\tilde{\bm{\theta}}^*)-\mathcal{F}(\bm{\theta}^*)\leq\frac{4}{m}|\tilde{\mathcal{V}}|C+\frac{L}{m}|\Delta\mathcal{V}|\cdot\|\bm{\theta}^*-\tilde{\bm{\theta}}^*\|.
\end{equation}
Next, according to \Cref{asp:convexity}, we have
\begin{equation}\label{eq:strongly_convex}
\mathcal{F}(\tilde{\bm{\theta}}^*)\geq\mathcal{F}(\bm{\theta}^*)+\nabla\mathcal{F}(\bm{\theta}^*)^\top(\tilde{\bm{\theta}}^*-\bm{\theta}^*)+\frac{\lambda}{2}\|\tilde{\bm{\theta}}^*-\bm{\theta}^*\|^2.
\end{equation}
Considering that $\bm{\theta}^*$ is a local optimum of $\mathcal{F}$, we have $\nabla\mathcal{F}(\bm{\theta}^*)=0$. Incorporate this condition into \Cref{eq:strongly_convex} and we have
\begin{equation}\label{eq:convex_deduction}
\begin{aligned}
\frac{\lambda}{2}\|\tilde{\bm{\theta}}^*-\bm{\theta}^*\|^2&\leq\mathcal{F}(\tilde{\bm{\theta}}^*)-\mathcal{F}(\bm{\theta}^*) \\
&\leq\frac{4}{m}|\tilde{\mathcal{V}}|C+\frac{L}{m}|\Delta\mathcal{V}|\cdot\|\bm{\theta}^*-\tilde{\bm{\theta}}^*\|.
\end{aligned}
\end{equation}
By telescoping \Cref{eq:convex_deduction}, we derive a quadratic function in terms of $\|\tilde{\bm{\theta}}^*-\bm{\theta}^*\|$:
\begin{equation}
\|\tilde{\bm{\theta}}^*-\bm{\theta}^*\|^2-\frac{2L|\Delta\mathcal{V}|}{m\lambda}\|\bm{\theta}^*-\tilde{\bm{\theta}}^*\|-\frac{8|\tilde{\mathcal{V}}|C}{m\lambda}\leq0.
\end{equation}
Finally, we finish the proof by solving this quadratic equation:
\begin{equation}
\|\tilde{\bm{\theta}}^*-\bm{\theta}^*\|\leq\frac{L|\Delta\mathcal{V}|+\sqrt{4m\lambda C|\tilde{\mathcal{V}}|+L^2|\Delta\mathcal{V}|^2}}{m\lambda}.
\end{equation}
\end{proof}

\begin{appendix_proposition}
\textbf{Distance Bound in Approximation.}
The $\ell_2$ distance bound between $\tilde{\bm{\theta}}^*$ and $\bar{\bm{\theta}}^*$ is given by 
\begin{align}
\|\tilde{\bm{\theta}}^*-\bar{\bm{\theta}}^*\|_2 \leq\frac{\lambda \|\Delta \bar{\bm{\theta}}^*\|_2 + L|\Delta\mathcal{V}|+\sqrt{4m\lambda C|\tilde{\mathcal{V}}|+L^2|\Delta\mathcal{V}|^2}}{m\lambda}.
\end{align}
\end{appendix_proposition}

\begin{proof}
Combine \Cref{influence} and $\bar{\bm{\theta}}^*=\bm{\theta}^*+\frac{1}{m}\Delta\bar{\bm{\theta}}^*$, we have
\begin{equation}
\begin{aligned}
\|\tilde{\bm{\theta}}^*-\bar{\bm{\theta}}^*\|_2&\leq\|\tilde{\bm{\theta}}^*-\bm{\theta}^*\|_2+\|\bm{\theta}^*-\bar{\bm{\theta}}^*\|_2 \\
&\leq\frac{L|\Delta\mathcal{V}|+\sqrt{4m\lambda C|\tilde{\mathcal{V}}|+L^2|\Delta\mathcal{V}|^2}}{m\lambda}+\frac{1}{m}\|\Delta\bar{\bm{\theta}}^*\|_2 \\
&=\frac{\lambda \|\Delta \bar{\bm{\theta}}^*\|_2 + L|\Delta\mathcal{V}|+\sqrt{4m\lambda C|\tilde{\mathcal{V}}|+L^2|\Delta\mathcal{V}|^2}}{m\lambda}.
\end{aligned}
\end{equation}
\end{proof}

\begin{appendix_theorem}
Let $\bm{\theta}^* = \mathcal{A}\left(\mathcal{G}\right)$ be the empirical minimizer over $\mathcal{G}$, $\tilde{\bm{\theta}}^* = \mathcal{A}\left(\mathcal{G} \ominus \Delta \mathcal{G}\right)$ be the empirical minimizer over $\mathcal{G} \ominus \Delta \mathcal{G}$ and $\bar{\bm{\theta}}^*$ be an approximation of $\tilde{\bm{\theta}}^*$.
Define $\zeta$ as an upper bound of $\|\tilde{\bm{\theta}}^*-\bar{\bm{\theta}}^*\|_2$. We have
$\mathcal{U}\left(\mathcal{G},\Delta\mathcal{G},\mathcal{A}\left(\mathcal{G}\right)\right) = \bar{\bm{\theta}}^*+\bm{b}$ is an ($\varepsilon-\delta$) certified unlearning process, where $\bm{b}\sim\mathcal{N}(0,\sigma^2\bm{I})$ and $\sigma\geq\frac{\zeta}{\varepsilon}\sqrt{2\mathrm{ln(1.25/\delta)}}$.
\end{appendix_theorem}

\begin{proof}
Given that $\zeta$ corresponds to the $\ell$-2 sensitivity of the approximating function, we can follow the same proof as the Theorem A.1 provided in \cite{dwork2014algorithmic} to obtain that $\mathcal{U}$ is $(\varepsilon,\delta)$ differentially private.
Consequently, our proof is finished following the Lemma 10 provided in \cite{sekhari2021remember} or the discussion of the relationship between certified unlearning and differential privacy provided in \cite{guo2019certified}.
\end{proof}

\begin{figure*}[t]
\centering
            \begin{subfigure}[t]{0.6\textwidth}
        \small
        \includegraphics[width=1.0\textwidth]{images/bound_legend.pdf}
        \end{subfigure} \\
        \begin{subfigure}[t]{0.33\textwidth}
        \small
        \centering
        \includegraphics[width=1.00\textwidth]{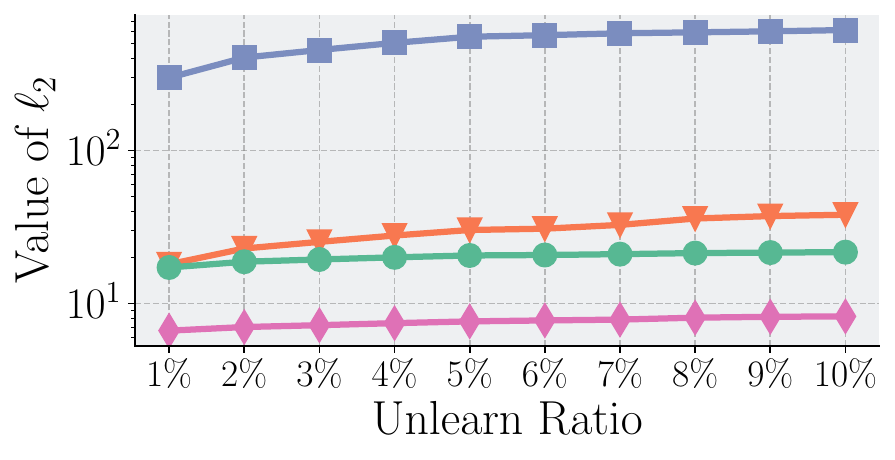}
            \caption[Network2]%
            {{Bounds vs. actual $\ell_2$ distance on Cora.}} 
            \label{cora_sgc}
        \end{subfigure}
            \begin{subfigure}[t]{0.33\textwidth}
        \small
        \centering
        \includegraphics[width=1.0\textwidth]{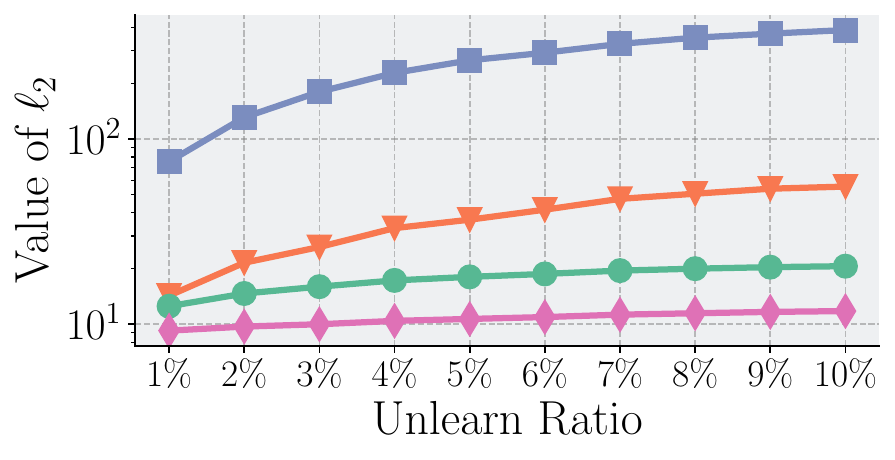}
            \caption[Network2]%
            {{Bounds vs. actual $\ell_2$ distance on CiteSeer.}}  
            \label{citeseer_sgc}
        \end{subfigure}
            \begin{subfigure}[t]{0.33\textwidth}
        \small
        \centering
        \includegraphics[width=1.0\textwidth]{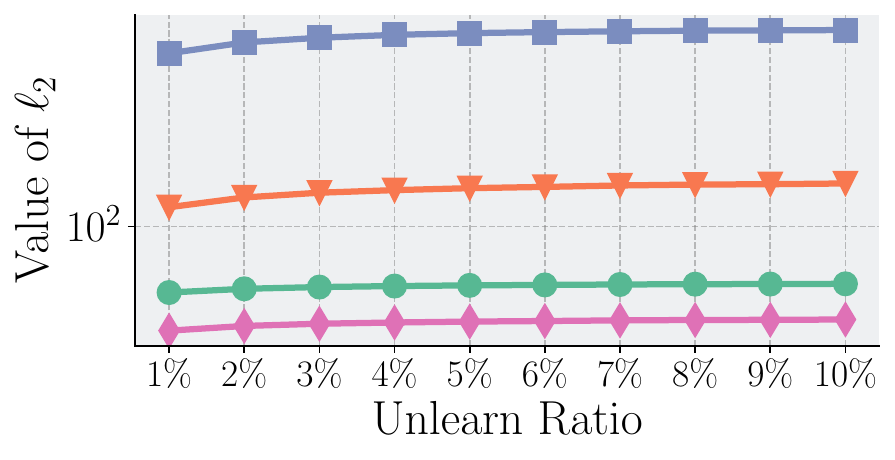}
            \caption[Network2]%
            {{Bounds vs. actual $\ell_2$ distance on PubMed.}}  
            \label{pubmed_sgc}
        \end{subfigure}
            \vspace{-3mm} 
    \caption{Bounds and actual value of the $\ell_2$ distance between $\tilde{\bm{\theta}}^*$ and $\bar{\bm{\theta}}^*$, i.e., $\|\tilde{\bm{\theta}}^*-\bar{\bm{\theta}}^*\|_2$, over Cora, CiteSeer and PubMed datasets. \textit{CEU Worst}, \textit{CEU Data Dependent}, \textit{IDEA}, and \textit{Actual} represent the worst bound based on CEU, the data-dependent bound based on CEU, the bound based on IDEA, and the actual value of $\|\tilde{\bm{\theta}}^*-\bar{\bm{\theta}}^*\|_2$ derived from re-training, respectively.}
    \label{sgc_bounds}
\end{figure*}

\begin{table*}[t]
\setlength{\tabcolsep}{27.65pt}
\renewcommand{\arraystretch}{0.8}
\centering
\vspace{-1mm}
\caption{The running time (in seconds) of edge unlearning on Coauthor-CS dataset based on GCN. CGU is excluded from comparison since its backbone only support SGC. The running time of the proposed IDEA is marked in bold.}
\label{unlearning_time_appendix}
\vspace{-3mm}
\begin{tabular}{lcccc}
\toprule
       & \textbf{0.1\%} & \textbf{1\%}  & \textbf{5\%} & \textbf{10\%} \\
       \midrule
\textbf{Re-Training}  &170.0 $\pm$ 4.5     & 170.7 $\pm$ 5.6         & 169.8 $\pm$ 7.1       & 201.2 $\pm$ 14         \\
\textbf{Random} & 11.54 $\pm$ 0.8     & 11.09 $\pm$ 0.2 & 11.20 $\pm$ 0.2     & 11.07 $\pm$ 0.5        \\
\textbf{BEKM}   &  9.66 $\pm$ 0.05  & 11.23 $\pm$ 0.3   & 10.88 $\pm$ 0.2     & 11.55 $\pm$ 0.3             \\
\textbf{BLPA}   & 10.75 $\pm$ 0.1     & 10.71 $\pm$ 0.3  & 10.77 $\pm$ 0.2     & 10.94 $\pm$ 0.2             \\
\textbf{IDEA}  & \textbf{2.823 $\pm$ 0.0}     & \textbf{2.874 $\pm$ 0.1}         & \textbf{2.851 $\pm$ 0.1}       & \textbf{3.328 $\pm$ 0.1}            \\
\bottomrule
\end{tabular}
\end{table*}

\begin{table*}[t]
\setlength{\tabcolsep}{8.65pt}
\renewcommand{\arraystretch}{0.8}
\centering
\vspace{-2mm}
\caption{Node classification accuracy after edge unlearning.  The results outside of brackets are accuracy values after unlearning with IDEA, while those inside brackets are given by re-training. The accuracy values of IDEA are marked in bold.}
\vspace{-3mm}
\label{unlearning_utility_appendix}
\begin{tabular}{lcccc}
\toprule
       & \textbf{0.1\%} & \textbf{1\%}  & \textbf{5\%} & \textbf{10\%} \\
       \midrule
\textbf{Cora} & \textbf{81.18 $\pm$ 0.9} ( 84.50 $\pm$ 0.3 )    & \textbf{81.18 $\pm$ 1.4} ( 84.38 $\pm$ 0.3 )         & \textbf{79.34 $\pm$ 0.8}   ( 82.41 $\pm$ 0.5 )     & \textbf{78.35 $\pm$ 1.0}  ( 82.29 $\pm$ 0.3 )         \\
\textbf{CiteSeer}   & \textbf{69.87 $\pm$ 1.3} ( 74.97 $\pm$ 0.3 )    & \textbf{69.87 $\pm$ 0.8} ( 75.38 $\pm$ 0.5 )         & \textbf{68.47 $\pm$ 0.8}   ( 72.97 $\pm$ 0.7 )     & \textbf{67.27 $\pm$ 0.6}  ( 73.57 $\pm$ 0.7 )          \\
\textbf{PubMed}   & \textbf{81.02 $\pm$ 1.0} ( 84.99 $\pm$ 0.1 )    & \textbf{80.90 $\pm$ 0.9} ( 84.77 $\pm$ 0.0 )         & \textbf{79.63 $\pm$ 0.9}   ( 83.82 $\pm$ 0.1 )     & \textbf{77.96 $\pm$ 0.8}  ( 81.91 $\pm$ 0.1 )    \\
\textbf{CS}   & \textbf{88.62 $\pm$ 5.4} ( 91.89 $\pm$ 1.6 )    & \textbf{92.42 $\pm$ 0.3} ( 92.73 $\pm$ 0.3 )         & \textbf{91.58 $\pm$ 0.3}   ( 92.82 $\pm$ 0.1 )     & \textbf{90.91 $\pm$ 0.7}  ( 91.84 $\pm$ 0.2 )    \\
\textbf{Physics}  & \textbf{95.68 $\pm$ 0.4} ( 96.11 $\pm$ 0.2 )    & \textbf{95.64 $\pm$ 0.1}  ( 96.03 $\pm$ 0.2 ) & \textbf{95.50 $\pm$ 0.3}  ( 96.00 $\pm$ 0.2 )     & \textbf{95.19 $\pm$ 0.3}  ( 95.86 $\pm$ 0.2 )       \\
\bottomrule
\end{tabular}
\end{table*}

\section{Supplementary Experiments}
\label{appendix_results}

\subsection{Evaluation of Bound Tightness}
In this subsection, we present additional experimental results regarding the bound tightness of the proposed model IDEA. Specifically, here we adopt SGC as our backbone GNN model, and we present the comparison between three bounds and actual value of the $\ell_2$ distance bound between $\tilde{\bm{\theta}}^*$ and $\bar{\bm{\theta}}^*$ in Figure~\ref{sgc_bounds}.

We present an review of the introduction for the bounds and the $\ell_2$ distances below (as in Section~\ref{bound_tightness_section}). 
\textit{(1) CEU Worst Bound.} We compute the theoretical worst bound derived based on CEU as a baseline of the $\ell_2$ distance bound between $\tilde{\bm{\theta}}^*$ and $\bar{\bm{\theta}}^*$.
\textit{(2) CEU Data-Dependent Bound.} We compute the data-dependent bound derived based on CEU as a baseline of the $\ell_2$ distance bound between $\tilde{\bm{\theta}}^*$ and $\bar{\bm{\theta}}^*$. Usually, data-dependent bound is tighter than the CEU Worst Bound.
\textit{(3) IDEA Bound.} We compute the bound given by Equation~\ref{final_triangle} as the IDEA bound for the $\ell_2$ distance bound between $\tilde{\bm{\theta}}^*$ and $\bar{\bm{\theta}}^*$. 
\textit{(4) Actual Values.} We compare the bounds above with the actual values of the $\ell_2$ distance between $\tilde{\bm{\theta}}^*$ and $\bar{\bm{\theta}}^*$.
In addition, we also follow the wide range of unlearning ratios as presented in Section~\ref{bound_tightness_section}.

Below we summarize the observations, which remains consistent with the observations presented in Section~\ref{bound_tightness_section} and are also found on other GNNs and datasets.
(1) From the perspective of the general tendency, we observe that when the value of unlearn ratio is increased (i.e., more edges are unlearned), it generally results in higher values for both the obtained bounds and the actual $\ell_2$ distance between $\tilde{\bm{\theta}}^*$ and $\bar{\bm{\theta}}^*$. This reveals that a higher unlearn ratio tends to make approximating $\tilde{\bm{\theta}}^*$ (using the calculated $\bar{\bm{\theta}}^*$) more difficult, which is also consistent with prior research~\cite{wu2023certified}.
(2) From the perspective of the bound tightness, the obtained results indicate that IDEA consistently provides tighter bounds than those produced by CEU across all scenarios, especially in cases with higher unlearn ratios. This suggests that the approximation of $\tilde{\bm{\theta}}^*$ offered by IDEA can capture the distance between $\tilde{\bm{\theta}}^*$ and $\bm{\theta}^*$ more accurately compared to CEU, especially for larger unlearn ratios.

\subsection{Evaluation of Unlearning Efficiency}

In this subsection, we present additional experimental results regarding the unlearning efficiency on different tasks. Specifically, we have shown efficiency comparison between IDEA and other baselines on node unlearning task in Section~\ref{efficiency_section}, and here we present additional results under edge unlearning task based on GCN, with all other settings being consistent with the experiments presented in Section~\ref{efficiency_section}. We show edge unlearning performance under a wide range of unlearning ratios (i.e., the ratio of edges to be unlearned to the total number of edges in the graph) from 0.1\% to 10\%. We present the experimental results in Table~\ref{unlearning_time_appendix}. We observe that the running time of the three variants of GraphEraser does not change as much as in Section~\ref{efficiency_section} across the wide range of ratios. This is because GraphEraser perform partition over the input graph, which results in different shards. Each shard will contribute to the overall running time when the unlearn node/edge appear in such a shard.
However, the number of edges is much more than the number of nodes. As a consequence, unlearn edges appears in most shards even if the unlearn ratio is as small as 0.1\%. Hence most shards contribute to the running time in most cases, leading to a more stable running time across all ratios.
Meanwhile, we also observe that IDEA not only achieves significantly less running time compared with the re-training approach, but also costs less running time compared with all other baselines. Such observation is consistent with the observation in Section~\ref{efficiency_section}, and can also be found on other datasets and GNNs. This indicates the superiority of IDEA in terms of the unlearning efficiency.

\begin{table}[H]
\setlength{\tabcolsep}{6.65pt}
\renewcommand{\arraystretch}{0.8}
\centering
\caption{AUC scores of attacks on Co-author CS based on GCN after node and edge unlearning, respectively. The results given by IDEA is marked in bold.}
\label{unlearning_effectiveness1_appendix}
\begin{tabular}{lcc}
\toprule
       & \textbf{Node Unlearning ($\downarrow$)} & \textbf{Edge Unlearning ($\downarrow$)}  \\
       \midrule
\textbf{Random} & 51.51 $\pm$ 0.7     & 50.26 $\pm$ 0.4        \\
\textbf{BEKM}   &  50.24 $\pm$ 0.8     & 50.26 $\pm$ 0.1               \\
\textbf{BLPA}   & 51.42 $\pm$ 0.8     & 50.04 $\pm$ 0.2              \\
\textbf{IDEA}   &  \textbf{50.15 $\pm$ 0.8}    &  \textbf{50.02 $\pm$ 0.7}            \\
\bottomrule
\end{tabular}
\end{table}

\begin{table}[H]
\setlength{\tabcolsep}{9.5pt}
\renewcommand{\arraystretch}{0.8}
\centering
\caption{Average loss values regarding the nodes with the unlearn attribute values being set to zero on CiteSeer. Lower values represents better unlearning performance, and the results given by IDEA are marked in bold.}
\vspace{-2mm}
\label{unlearning_effectiveness2_appendix}
\begin{tabular}{lccc}
\toprule
       & \textbf{20\% ($\downarrow$)} & \textbf{50\% ($\downarrow$)} & \textbf{80\% ($\downarrow$)} \\
       \midrule
\textbf{Random} & 1.44 $\pm$ 0.09     & 1.43 $\pm$ 0.08     & 1.48 $\pm$ 0.12          \\
\textbf{BEKM}   &  1.44 $\pm$ 0.06   & 1.50 $\pm$ 0.04    & 1.51 $\pm$ 0.09               \\
\textbf{BLPA}   &  1.40 $\pm$ 0.19   &  1.45 $\pm$ 0.27   & 1.50 $\pm$ 0.09              \\
\textbf{IDEA}    & \textbf{1.25 $\pm$ 0.01}    & \textbf{1.28 $\pm$ 0.01}    & \textbf{1.33 $\pm$ 0.02}             \\
\bottomrule
\end{tabular}
\end{table}

\subsection{Evaluation of Model Utility}

We now present additional results for the evaluation of model utility. We have shown the node classification accuracy comparison based on the SGC model in Section~\ref{utility_section}, and here we show the node classification accuracy comparison between IDEA and re-training based on the GCN model.
We maintain all other settings to be consistent with those in Section~\ref{utility_section}. The model utility given by IDEA and re-training is compared across a wide range of unlearn ratio values (from 0.1\% to 10\% ). We observe that the unlearning given by IDEA only sacrifices limited utility compared with re-training, which remains consistent with the observation in Section~\ref{utility_section}, and such an observation can also be found on other datasets, unlearning tasks and GNNs. This indicates the satisfying usability of IDEA.

\subsection{Evaluation of Unlearning Effectiveness}

Finally, we present additional results for the evaluation of unlearning effectiveness. To evaluate the generalization capability of IDEA, here we utilize a different GNN model (compared with the results in Section~\ref{effectiveness_section}), which is GCN, as the backbone. We adopt a consistent evaluation protocol as shown in Section~\ref{effectiveness_section}, and we present the experimental results of node/edge unlearning and attribute unlearning in Table~\ref{unlearning_effectiveness1_appendix} and Table~\ref{unlearning_effectiveness2_appendix}, respectively. We found that, first, in terms of node and edge unlearning tasks, we observe that the attacks on IDEA show the lowest attack successful AUC scores, which are marginally above 50\% (almost equivalent to random guess). This indicates the effectiveness of IDEA in performing node and edge unlearning.
Second, we observe that the average loss values given by IDEA are the lowest among all baselines. Since the loss values are collected from the nodes whose unlearn attribute values have already been set to zero, a lower loss value indicates better unlearning effectiveness. Therefore, the satisfying effectiveness of IDEA is further validated. Additionally, we note that these observations can also be found on other unlearning tasks, datasets, and GNNs, which indicates the satisfying unlearning effectiveness of IDEA.

\end{document}